\RequirePackage{snapshot}
\documentclass[twoside]{article}

\usepackage[accepted]{aistats2022}
\usepackage{custom}
\usepackage[round]{natbib}

\begin{document}

\runningtitle{PAC$^m$-Bayes}

%

%

\twocolumn[

\aistatstitle{PAC$^m$-Bayes \\ {\small Narrowing the Empirical Risk Gap in the Misspecified Bayesian Regime}}
\aistatsauthor{ Warren R.~Morningstar \And Alexander A.~Alemi \And  Joshua V.~Dillon }
\aistatsaddress{ Google Research \And Google Research \And Google Research } ]

\begin{abstract}

The Bayesian posterior minimizes the ``inferential risk'' which itself bounds the ``predictive risk.'' This bound is tight when the likelihood and prior are well-specified. However since misspecification induces a gap, the Bayesian posterior predictive distribution may have poor generalization performance.  This work develops a multi-sample loss (\pacm) which can close the gap by spanning a trade-off between the two risks. The loss is computationally favorable and offers PAC generalization guarantees. Empirical study demonstrates improvement to the predictive distribution.

\end{abstract}

%


\section{INTRODUCTION}

\begin{figure}[htb]
  \centering
  \includegraphics[width=\linewidth]{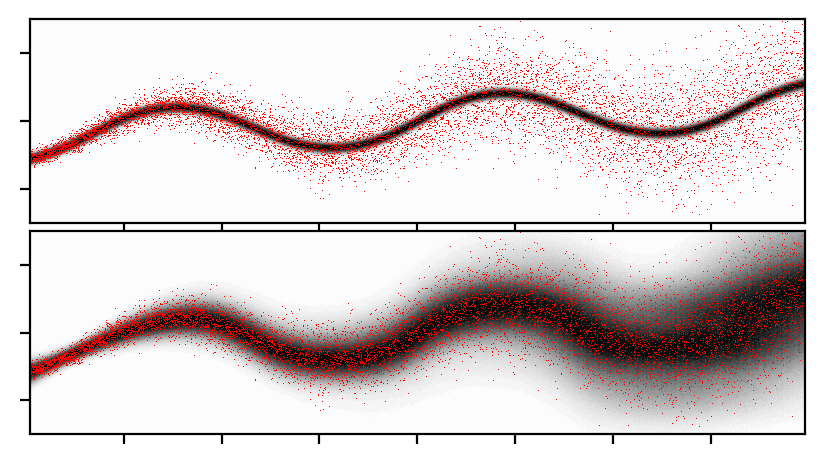}
  \caption{\label{fig:hero}\footnotesize{A misspecified Bayesian neural network changes from bad uncertainty estimation in the posterior predictive distribution (top row), to good uncertainty estimation in the (approximate) posterior predictive distribution (bottom row) with a simple (one line) change to the training loss. 
  Cutout from~\cref{fig:toy_experiments}.}}
\end{figure}

The top and bottom of~\cref{fig:hero} differ by one line of code. 
The traditionally inferred (approximate) posterior (top) fails to capture heteroskedastic
noise in the data while the proposed generalization (bottom) succeeds. 
Both rows employ the same data, computation, model family, and optimization procedure; 
only the loss differs in that the first row is based on a average-log-likelihood and the second is based on a log-average-likelihood.
%
Before we can understand how/why this works and return to this example (\cref{sec:experiments,fig:toy_experiments}), we have to examine the difference between prediction and inference.

Pierre-Simon Laplace formulated one of the earliest Bayesian models~\citep{laplace1781memoire}. Interested in the relative birth rates of boys and girls, 
he derived the Beta posterior for a Bernoulli likelihood with uniform prior.
He then calculated the ``posterior probability'' that the girl birth rate exceeds the boy rate and found it to be about $10^{-42}$ from which he concluded that it is ``as certain as any other moral truth'' that humans give birth to more boys than girls.\footnote{The presently accepted natural ratio is 105 males per 100 females~\citep{owidgenderratio}}

Laplace's objective was to \emph{infer} the parameter of his model.
Broadly, science has followed suit.  When a modern experiment such as the Large Hadron Collider at CERN processes terabytes of particle collision data~\citep{atlas2012}, or the Planck satellite maps the cosmic microwave background radiation~\citep{planck2018}, they are in pursuit of the ``moral truth'' of some underlying parameter of the universe.

Contrast this with modern machine learning.
The primary goal of machine learning
is to build models that can form
accurate \emph{predictions}. 
We do 
not truly care about the value of the millionth weight in a deep neural network.  We do not believe the parameters of the 
neural network are reflecting any ``moral truths''.

For well-specified models the goals of inference and prediction align.
For misspecified models they might not.
Optimizing for inference when you'll evaluate a model's predictive performance can thus lead to sub-optimal predictive models.
As made clear in the work of~\citet{masegosa2019learning}, both Bayesian inference and Maximum
Likelihood target inferential rather than predictive risks, and
can make poor predictions under model misspecification.

This issue is well known~\citep{minka2000bayesian,domingos1997does},
and much prior work has been done to address it~(e.g.~\citep{berger1994overview,yao2018using,bissiri2016general,grunwald2017inconsistency,jankowiak2020deep,jankowiak2020parametric,masegosa2019learning,wei2020direct,pmlr-v118-sheth20a}).

In this work, we introduce a tractable multi-sample bound on the true predictive risk which sometimes manifests in the striking improvement demonstrated in in \cref{fig:hero,fig:toy_experiments}. This bound interchanges average and log and enables recovering ordinary Maximum Likelihood and the Bayesian posterior.  We list our contributions as:
\begin{enumerate}
    \item We introduce a novel multisample bound on the true predictive risk, which we call \pacm-Bayes.
    \item We show that this bound can be a tighter bound on the predictive risk than similar bounds on the inferential risk, which are widely used in practice.
    \item We prove that the slack of this bound is bounded under similar sets of assumptions as used in other PAC-Bayes works.
    \item We present empirical study demonstrating that \pacm-Bayes leads to models which better approximate the true predictive distribution than alternative bounds on the predictive risks.
\end{enumerate}


\section{PREDICTIVE AND INFERENTIAL RISKS}


We begin at a high level 
with a \emph{statistical model}: $p(X| \theta)$ defining a distribution of each observed datum
$X$ in terms of some parameters $\theta$.
After observing $n$ data points drawn from some \emph{true distribution} $X^n\defeq\{X_i\}_i^n\iid\nu(X)$, we form a distribution of parameters, $q(\Theta|\{x_i\}_i^n)$.  
In principle we can then compute the \emph{predictive distribution}:
\begin{equation}
    \label{eqn:predictive}
    p(X| \{x_i\}_i^n) = \E_{q(\Theta|\{x_i\}_i^n)} \left[ p(X|\Theta) \right].
\end{equation}
(For brevity we henceforth regard $q$'s dependence on $\{x_i\}_i^n$ as implicit.)
If we had some particular application in mind, at this point we could score our model's ability to make predictions as measured by some specific \emph{risk}, a path that would lead to the general field of Bayesian risk minimization~\citep{berger2013statistical}.
To keep things simple here, lacking a specific risk, 
we judge the quality of our predictive distribution by measuring the relative entropy (Kullback-Leibler divergence) between the true distribution and our predicted one:
\begin{multline}
    \KL[\nu(X); p(X|q)] = \E_{\nu(X)}\left[ \log \frac{\nu(X)}{p(X|q)} \right] \\
    = \E_{\nu(X)}[\log \nu(X)]
    -\E_{\nu(X)}\left[ \log p(X|q) \right].
\end{multline}
Up to a constant outside our control (the continuous entropy of the true distribution) this defines what we'll call the \emph{true predictive risk}:
\begin{equation}
    \label{eqn:truepredrisk}
    \truepredrisk[q] \defeq -\E_{\nu(X)}\left[ \log \E_{q(\Theta)}[p(X|\Theta)] \right]. 
\end{equation}

In many cases the true predictive risk is ultimately what we care most about. Determining how accurately we can predict the future, it is often what governs how much money our model will make or how many lives it will save. 

Not knowing the true distribution $\nu(X)$, 
we cannot directly minimize the true predictive risk. 
One thing we can compute is the \emph{empirical predictive risk}:
\begin{equation}
    \label{eqn:emppredrisk}
    \emppredrisk_n[q] \defeq -\frac 1 n \sum_i^n \log \E_{q(\Theta)}[p(x_i|\Theta)].
\end{equation}
This is the observed average risk on the $\{x_i\}_i^n$ sample set. 
Akin to a 
training loss, the empirical predictive risk is a measure
of how well we do at predicting the training data.
If used as a target
for optimization we can easily overfit.  Training with this risk directly would amount to a type of \emph{ensemble method}~\citep{dietterich2000ensemble} or \emph{non-parametric mixture} with mixing distribution $q(\Theta)$~\citep{wangnonparam,lindsay1995mixture}.

In contrast to the predictive risks, we'll also define the \emph{inferential risks} which focus on determining or inferring the correct values of the parameters.
The \emph{true inferential risk} (often just called the \emph{true risk}):
\begin{equation}
    \label{eqn:trueinfrisk}
    \trueinfrisk[q] \defeq -\E_{\nu(X)}\left[ \E_{q(\Theta)} \left[ \log p(X|\Theta)  \right]\right],
\end{equation}
and the corresponding \emph{empirical inferential risk} (often called the \emph{empirical risk}):
\begin{equation}
    \label{eqn:empinfrisk}
    \empinfrisk_n[q] \defeq -\frac 1 n \sum_i^n \E_{q(\Theta)} \left[ \log p(x_i|\Theta)  \right].
\end{equation}

For a variety of reasons, directly minimizing the inferential risk is fairly commonplace. It measures the average of the divergence between the true distribution $\nu(X)$ and
the single-value parameter settings of the model $p(X|\theta)$. 
This is akin to doing variational optimization~\citep{varopt}, 
and concentrates on a delta function corresponding to the best
single-value parameter setting. 

Jensen's inequality implies 
\[ -\log \E_{q(\Theta)}[p(x|\Theta)] \leq -\E_{q(\Theta)}[\log p(x|\Theta)] \]
and so the inferential risks are upper bounds on the  predictive risks:
\begin{equation}
    \truepredrisk[q] \leq \trueinfrisk[q] \qquad \text{ and } \qquad\emppredrisk_n[q] \leq \empinfrisk_n[q].
\end{equation}

In this way, minimizing the inferential risk is a \emph{valid} strategy for achieving good predictions since it minimizes an upper bound on the predictive risk.  
When is it a \emph{good} strategy?  When is this bound tight? Answer: If our model is well-specified~\citep{masegosa2019learning} (Proof replicated in~\cref{sec:wellspecifiedtight}).
However, in cases of model misspecification this can break down severely.
We would prefer to target the true predictive risk directly,
but cannot since we do not know the true data distribution.
What we need is a tractable bound on the true risks.


\section[PAC-BAYES]{\pac-BAYES}

While the empirical risks ($\emppredrisk, \empinfrisk$) provide
unbiased estimates of the true risks ($\truepredrisk, \trueinfrisk$),
minimizing the empirical risks do not minimize the true risks:
\begin{equation*}
    \argmin_q \trueinfrisk[q] = \argmin_q \E\left[ \empinfrisk_n[q] \right] \neq \E\left[ \argmin_q \empinfrisk_n[q]  \right].
\end{equation*}
Said another way, the empirical risks do not provide a \emph{bound} on the true risks.

Despite not being a valid bound, empirical risk minimization is quite popular.  Minimizing the empirical (inferential) risk over the space of all possible distributions over parameters is the well known Maximum Likelihood method.
This concentrates in a delta-function-like parameter distribution
with all of its mass on the maximum likelihood parameter value. 

We could similarly directly optimize the empirical predictive risk, known to some as a \emph{non-parametric mixture}~\citep{lindsay1995mixture,wangnonparam}. In cases with bounded likelihoods this seems to perform decently well (e.g. the toy example of~\cref{sec:toy}) just as it does in the case of Maximum
Likelihood. 
If our model is too expressive minimizing the empirical risks 
will quickly start to concentrate 
on the \emph{empirical} data distribution
rather than the \emph{true} distribution, overfitting severely.
Classic approaches prevent overfitting by limiting model capacity; by adding regularization
or other tricks. 
If we instead had a valid bound on the true risks,
we needn't worry.
\pac-Bayes approaches provide such a bound.

We would really like to have some assurance that we 
won't overfit to our finite training data.
We can formulate an upper bound on the true risks
in terms of the empirical risks that nearly always hold.
Such \emph{probably approximately correct} (or \pac) bounds can be used to motivate Bayesian inference, demonstrating that the Bayesian posterior is the minimizer of a \pac-style upper bound on the true inferential risk $\trueinfrisk$~\citep{banerjee2006bayesian, alquier2016properties,guedj2019primer} (Proof replicated in~\cref{sec:pacbayesproof}).

In light of these results we will define the following \emph{\pac-inferential risk} (or \elbo):
\begin{multline}
    \label{eqn:pacinfrisk}
   \pacinfrisk_n[q; r, \beta] \defeq \empinfrisk_n[q] + \frac{1}{\beta n } \kl{q(\Theta)}{r(\Theta)}  \\
   = \E_{q(\Theta)} \left[ -\frac 1 n \sum_i \log p(x_i| \Theta) + \frac{1}{\beta n } \log \frac{q(\Theta)}{r(\Theta)}  \right].
\end{multline}
Aside from constants independent of $q$, $\pacinfrisk$ is a stochastic upper bound on $\trueinfrisk$.  Intuitively, this is accomplished 
by ensuring that our parameter distribution $q(\Theta)$ can't stray
too far from a \emph{prior} $r(\Theta)$ we chose before looking 
at the data.
Notice that ordinary Bayesian inference corresponds to minimizing this risk for $\beta = 1$~\citep{knoblauch2019generalized,bissiri2016general}. 
Furthermore, as $\beta \to \infty$ we recover the empirical risk $\empinfrisk$ and thus Maximum Likelihood.  This risk is well known and widely used, both from previous work on information theoretic bounds for statistical explanation \citep{zhang2006information, alemi2016deep}, and as the evidence lower bound (ELBO) from work on Variational Inference such as \cite{kingma2013auto}.

Because $\truepredrisk \leq \trueinfrisk$, Bayesian inference is equivalent to (almost always) minimizing an upper bound on the \emph{true predictive risk} $\truepredrisk$.  In the case of a well-specified model, $\min \truepredrisk = \min \trueinfrisk$ and Bayesian inference targets
not only optimal inferential power but also optimal predictive power.
If you have the correct model, searching for the correct single parameter setting of the model is the right thing to do.  Is this still the case when the model is misspecified?

We adopt the definition of model misspecification used in \cite{masegosa2019learning}, namely that the true data generating distribution is not recoverable using a single parameter setting of the predictive model ($\nu \not \in \{\ell(\cdot|\theta) : \theta \in\mathcal{T}\}$).  If the true data generating distribution is not measurable using a single parameter setting of our model, then as the distribution over parameters concentrates (as would happen when minimizing $\empinfrisk$ or $\pacinfrisk$ with infinite data) we cannot recover a perfect approximation of the true data generating distribution.  Therefore, if you have a misspecified model, searching for the best single parameter setting of that model is not the right thing to do.

What ought we do if our model is misspecified?

\section[PACm-BAYES]{\pacm-BAYES}

If our model is misspecified, there may be a large gap between the minimum of the predictive and inferential risks ($\min \truepredrisk \ll \min \trueinfrisk$) as we'll demonstrate in our experiments below.

Our central contribution is to provide a new class of bounds, analogous to the \pac-style upper bounds on the inferential risk but targeting the predictive risk more directly.

The potential gap between
the predictive and inferential risks came from invoking
Jensen's inequality:
\begin{equation}
    -\log \mathbb{E}_{q(\Theta)}[ p(x|\Theta) ] \leq -\mathbb{E}_{q(\Theta)}[ \log p(x|\Theta) ].
\end{equation}

The core insight is to explore a family of multisample stochastic bounds:~\citep{burda2015importance,mnih2016variational}
\begin{multline}
    -\log \mathbb{E}_{q(\Theta^m)}[ p(x|\Theta^m) ] \\
    \le -\mathbb{E}_{q(\Theta^m)}\left[ \log  \frac 1 m \sum_{j}^m p(x|\Theta_j)   \right] \\
    \le -\mathbb{E}_{q(\Theta)}[ \log p(x|\Theta) ].
\end{multline}
Averaging a finite number of samples from our parameter distribution provides an unbiased estimate of the predictive likelihood.  Taking the log of an unbiased estimator produces a \emph{stochastic lower bound}~\citep{burda2014accurate,grosse2016} that becomes tight asymptotically. 

\begin{figure*}[htbp]
\begin{theorem}
\label{thm:pacm}
        For all $q(\Theta)$ absolutely continuous with respect to $r(\Theta)$, $X^n\iid \nu(X)$, $\beta \in (0,\infty)$, $n,m\in \mathbb{N}$, $p(x|\theta) \in (0,\infty)$ for all $\{x \in \mathcal{X}:\nu(x)>0\} \times \{\theta \in \mathcal{T}: r(\theta)>0\}$, and $\xi \in (0,1),$ then with probability at least $1-\xi,$
    \begin{equation}
        \truepredrisk[q] \leq \pacpredrisk_{m,n}[q;r,\beta]  
        + \psi(\nu, \beta, m, n, r, \xi)
        - \tfrac{1}{\beta m n} \log \xi
    \end{equation}
    and furthermore (unconditionally),
    \begin{equation} 
        \pacpredrisk_{m,n}[q;r,\beta] 
        \leq \pacpredrisk_{m-1,n}[q;r,\beta]
        \leq \pacpredrisk_{1,n}[q;r,\beta]
        =   \pacinfrisk_n[q,r,\beta] 
         \label{eqn:thm1-2}
    \end{equation}
    where:
    \begin{align}
    \pacpredrisk_{m,n}[q;r,\beta] 
      &\defeq
       - \frac{1}{n} \sum_i^n \E_{q(\Theta^m)}\left[  \log \left( \frac{1}{m} \sum_j^m p(x_i | \Theta_j) \right) \right] + \frac{1}{\beta n} \kl{q(\Theta)}{r(\Theta)}
       \defeq \textrm{\pacm}  \label{eqn:pacpredrisk} \\
    \psi(\nu, \beta, m, n, r, \xi) &\defeq
      \tfrac{1}{\beta m n} \log \E_{\nu(X^n)}  \E_{r(\Theta^m)} \left[ e^{\beta n m \Delta(X^n, \Theta^m)}  \right]  \label{eqn:psi} \\
    \begin{split}
        \Delta(X^n, \Theta^m) 
        &\defeq
      \frac{1}{n}\sum_i^n \log \left( \frac{1}{m} \sum_j^m p(X_i|\Theta_j) \right) -\E_{\nu(X)}\left[\log \left( \frac{1}{m} \sum_j^m  p(X|\Theta_j) \right) \right].
     \end{split}
       \label{eqn:gap}
    \end{align}

\begin{proof}
Proof in~\cref{app:pacmproof}. \emph{Sketch}: form a multisample bound on the predictive risk and apply the traditional \pac-Bayes bound.
\end{proof}
\end{theorem}
\end{figure*}

Our main result is in~\cref{thm:pacm} (below) and \cref{thm:psi_bound} (appendix).
This \pac-Bound establishes that we are free to minimize the empirical predictive risk for any finite $m$, without fear of overfitting, provided we simultaneously ensure that our parameter distribution
remains close to some \emph{prior} $r(\Theta)$ which we specified independent of the data and which offers a reasonable explanation of our prior beliefs as to the model parameters in the absence of evidence.  
This is nearly always an upper bound on the true risk with a gap $\psi$ (\cref{eqn:psi}), a term which measures the discrepancy  
between true and empirical inferential risks ($\Delta$,~\cref{eqn:gap}) if we drew
parameter values from our prior.
Crucially, $\psi$ is independent of $q(\Theta)$ and can be disregarded from optimization.
We further show (\Cref{thm:psi_bound}, \Cref{app:pacmproof}) that under certain assumptions, $\psi$ is bounded and therefore \Cref{thm:pacm} is non-vacuous. 
If $\beta_{nm} = O(1)$ then $\psi_n = o(n).$ Because computational complexity increases with $m,$ asymptotic analysis of $m$ is not relevant. Nevertheless were we to consider large $m$, then $\psi_{m,n}=O(m)$ when $\beta_{nm}=O(m^{-1})$ and at best, $\psi_{m,n}=O(\log m)$ when $\beta_{nm}=O(m^{-1}\sqrt{\log m}).$ 
For more discussion and analysis see \cref{app:pacmproof,sec:vacuous}.

This yields our proposed risk, $\pacpredrisk_{n,m}$ (\cref{eqn:pacpredrisk}).
Minimizing $\pacpredrisk_{n,m}$~(\cref{eqn:pacpredrisk}) is equivalent to minimizing a stochastic upper bound on the true predictive risk $\truepredrisk$, analogous to the relationship between
$\pacinfrisk$  and $\trueinfrisk$.
See~\cref{thm:pacmproof} for a complete proof, though it follows directly
from the traditional PAC-Bayes proof once we invoke the multisample bound. 
Furthermore, as we increase $m$, \pacm\ decreases~(\cref{eqn:thm1-2}).

Dropping $\psi$ (being constant in $q$, though there is a lot of nuance here, see~\cref{app:pacmproof,sec:vacuous}), we can summarize the relationships between the risks as:
\begin{equation}
    \truepredrisk \lesssim \pacpredrisk_{n,m}
    \leq \pacpredrisk_{n,1} = \pacinfrisk_n 
    \gtrsim \trueinfrisk \geq \truepredrisk.
\end{equation}
The $\pacinfrisk_n \gtrsim \trueinfrisk$ relationship  is the classic \pac-Bayes result~\citep{alquier2016properties}, 
$\trueinfrisk \geq \truepredrisk$ follows from Jensen's inequality~\citep{masegosa2019learning}, and
the left hand side $\truepredrisk \lesssim \pacpredrisk_{n,m} \leq \pacpredrisk_{n,1} = \pacinfrisk_n$ is our contribution. 

\citet{masegosa2019learning} identified the need for tighter 
bounds on predictive risks than $\trueinfrisk$, suggesting
a family (\pactwot) of risks that utilize a second order Jensen bound.
While estimating the variance term in \pactwot\ 
requires care, \pacm\ is minibatch friendly, having its
expectation over the parameter distribution outermost in the
objective.
Later in our experiments we directly compare these approaches.

We now have two knobs we can use to adjust our risk: $m$, the number of samples we use to estimate the predictive distribution and $\beta$, a sort of inverse temperature used for adjusting the relative strength of the likelihood and prior terms.  For $m=1$ we 
recover the (inferential) risks we are used to, but for $m \ge 1$ we may form tighter bounds on the true predictive risk.  With $\beta=1$ we recover traditional Bayesian inference with an equal weighting of the likelihood and prior terms, as $\beta \to \infty$ we recover purely empirical risks.  For any $\beta$, including $\beta \le 1$ we still maintain our stochastic bounds.  Downweighting the $\KL$ term with respect to the prior, or \emph{cold posteriors} has shown to be useful especially in the context of neural networks~\citep{wenzel2020good}.

What do realizations of \pacm\ look like in practice?  In practice it amounts to a very simple change to existing variational Bayesian approaches. 
As can be seen in~\cref{fig:elbo-code,fig:pacm-code}, this can be a one line change, changing an expectation over draws from the variational posterior with a $\textsf{logsumexp}$.  Instead of scoring the average log likelihood of $m$ draws from a variational posterior, we instead score the log of the average likelihood across $m$ draws from a variational posterior.

\section{AN ILLUSTRATIVE TOY EXAMPLE}
\label{sec:toy}

Consider trying to fit a \normal\ distribution to a set of observations with a fixed unit variance but unknown mean:
\begin{equation}\label{eq:toylik}
    p(x|\theta) = \textrm{\normal}(x; \theta, 1) = (2\pi)^{-\frac{1}{2}} e^{-\frac{(x-\theta)^2}{2}} .
\end{equation}

Imagine further that we are operating in a severe model misspecification regime.  While our model is a unit variance \normal\ distribution, the true data distribution is a 30-70 mixture of two \normal s with twice the standard deviation and separated by four times their standard deviation.

\begin{figure}[hbt]
    \includegraphics[width=0.98\linewidth]{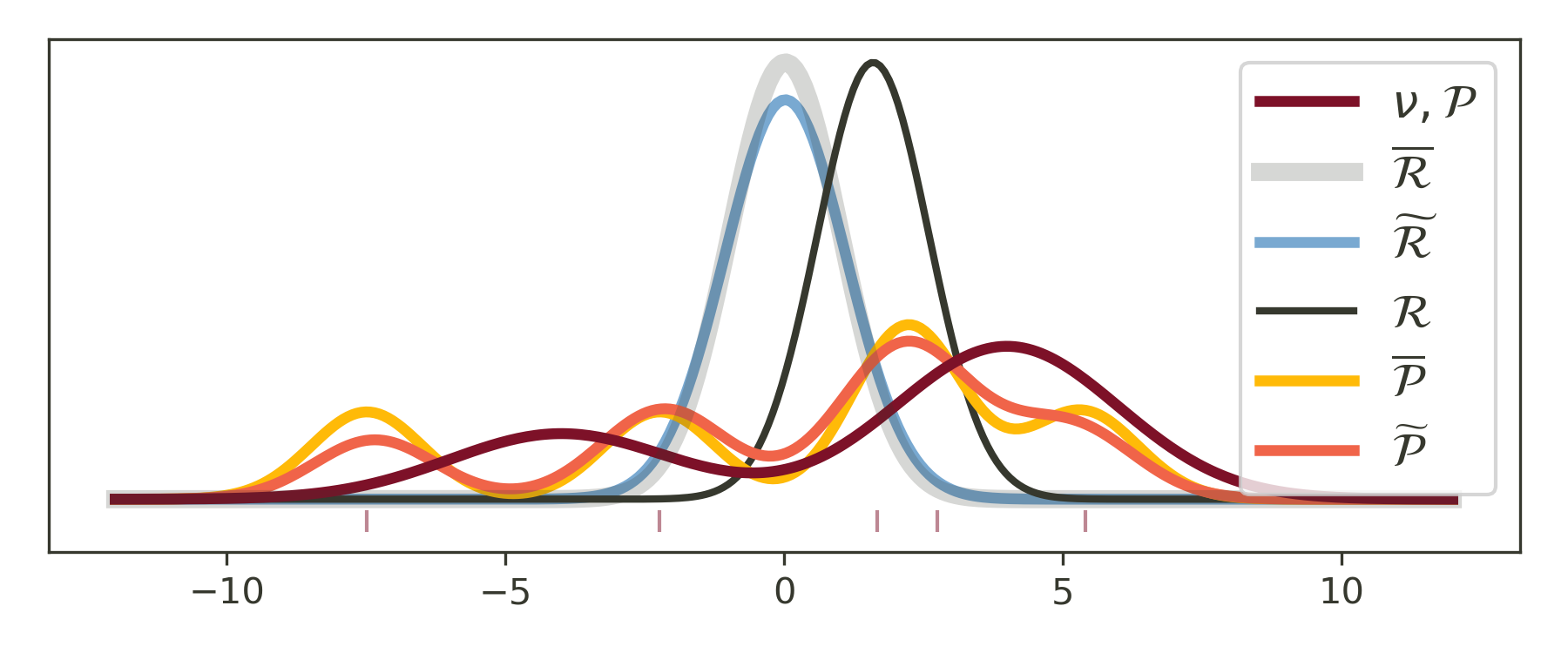} 
    \includegraphics[width=0.98\linewidth]{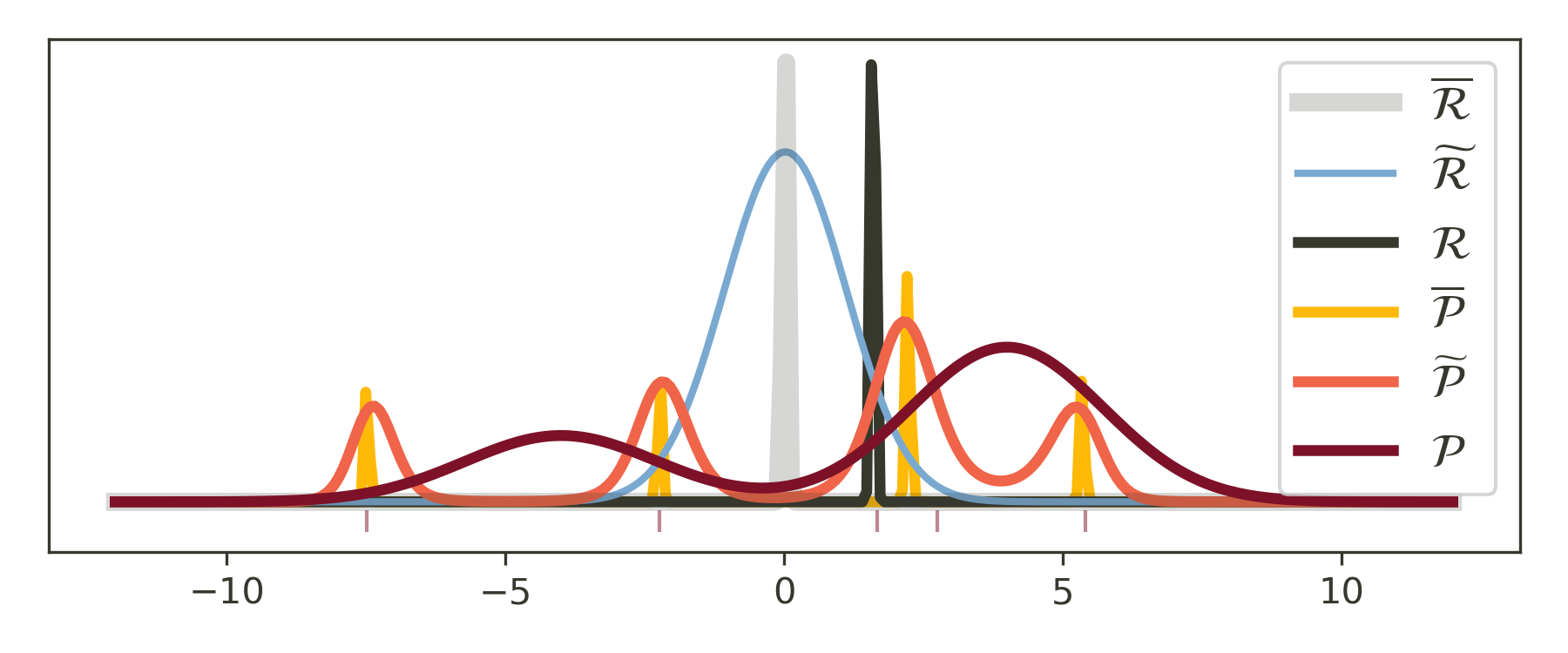}
    \caption{\footnotesize Toy Example: The top plot shows
    the resulting predictive distributions. The 
    bottom plot shows the learned parameter distributions.
    Please see accompanying text for a full explanation.
    Measured in bits, the KL divergences between the true distribution and each method are: $\empinfrisk: 12., \pacinfrisk: 9.6, \trueinfrisk: 10.0, \emppredrisk: 0.5, \pacpredrisk: 0.38, \truepredrisk: 0.0$
    } \label{fig:toy}
\end{figure}

In~\cref{fig:toy} (left) we show the predictive distributions that result from minimizing all of the risks discussed previously.  In~\cref{fig:toy} (right) we show the corresponding parameter distributions.
The true data distribution is shown with the dark red curve in~\cref{fig:toy} (left). 
The dark red tick marks on the axis show five ($n=5$) samples which we took as our data. The \cref{fig:toy} caption lists the resulting $\KL$ divergences between the true data distribution and each of the found predictive distributions.

Minimizing $\empinfrisk$ (\cref{eqn:empinfrisk}) is equivalent to Maximum Likelihood (grey curves),
which concentrates its parameter distribution to a delta-function located
at the empirical mean, and whose predictive distribution is simply a unit
variance \normal\ distribution centered at that empirical mean.  

Minimizing
$\pacinfrisk$ (\cref{eqn:pacinfrisk}, aka \elbo) is equivalent to Bayesian inference (blue curves). 
Here, we used a weakly informative $\textrm{\normal}(0,9^2)$ prior. 
This risk prevents the parameter distribution from collapsing onto a delta function, but the resulting predictive distribution is quite similar to the
one we found with Maximum Likelihood.  It is still fundamentally unimodal as
minimizing $\pacinfrisk$ is still fundamentally looking for the best single
parameter setting of our model. 

Minimizing $\trueinfrisk$ (\cref{eqn:trueinfrisk}) is equivalent to 
Bayesian inference with infinite data (black curve).  Here again our 
parameter distribution concentrates on a delta function, this time at the
true distributions mean, but the resulting predictive distribution is 
the best single parameter setting we could achieve, a unimodal predictive distribution that doesn't match the true distribution all that well. Notably, in this case it gives worse predictions than $\pacinfrisk$.

In contrast, minimizing the predictive risks (warm colors) do not look for
single parameter settings of the model.  Minimizing $\emppredrisk$
(\cref{eqn:emppredrisk})
performs
a sort of clustering of the data (yellow curve).  While it might seem natural 
to allow each data point its own delta-like contribution in the parameter 
distribution, two of our samples are near enough that we achieve better
empirical predictive risk by combining the two points into a single 
contribution to the parameter distribution with twice the weight
but located at the two points' mean. 
The resulting predictive distribution remains multimodal and achieves a much lower divergence with respect to the true distribution (0.5 bits versus the $\sim 10$ bits for the traditional (inferential) risks).  

Minimizing $\pacpredrisk$ (\cref{eqn:pacpredrisk}, aka \pacm, here with $m \to \infty$ see~\cref{sec:toydetails})
has  a similar qualitative effect compared to the corresponding inferential case ($\pacinfrisk$). 
The addition of the $\KL$ penalty with respect to some prior (here the same as used in the Bayesian case) prevents the parameter distribution from collapsing to a delta-comb.  

Finally, in this case, even though we have rather gross model misspecification in the sense that our model $p(X|\Theta)$ is quite unlike the true distribution for any single value of $\Theta$, our true distribution can be expressed as an infinite mixture of our model.
Minimizing the true $\truepredrisk$ (\cref{eqn:truepredrisk}) can achieve
perfect predictive performance (red curve).
This is achieved with a bimodal \normal\ distribution in parameter space which when convolved with our \normal\ model gives the exact bimodal \normal\ data distribution we chose.  This is also what we achieve asymptotically from $\pacpredrisk$ in the limit of infinite data.

This toy example illustrates how and when we can hope to achieve better predictive performance from $\emppredrisk,\pacpredrisk$ than from $\empinfrisk,\pacinfrisk$.  
Namely, if some mixture of our model can get closer to the true distribution than the best single setting of the parameters, we expect approaches that target the predictive risks to outperform the inferential risks by a corresponding margin.  

\section{RELATED WORK}



The work most closely related to the \pacm\ bound is the \pactwot\ bound presented in~\citet{masegosa2019learning}.  
\pactwot\ is based on a second order Jensen tightening of $\truepredrisk$. While clearly instrumental to our work, \pactwot\ has a number of defects which \pacm\ remedies. 
First, the variance tightening term in \pactwot\ is non-degenerate 
only for bounded likelihoods;
\pacm\ has no such restriction. 
Second, the \pacm\ risk, 
by directly targeting predictive risk
satisfies the \emph{golden rule};
the same cannot be said for \pactwot. 
Finally, in the experiments below we demonstrate that \pacm\ generally matches or exceeds the test-set performance of \pactwot;
as expected, both generally outperform \elbo.


The \pacm\ proofs leverage multisample insights from the IWAE work \citep{burda2015importance}. 
In response, \citet{rainforth2018tighter} question the utility of these tighter class of bounds and demonstrate that tighter bounds on the \emph{marginal evidence} do not help learn useful posteriors. 
This valuable insight does not apply to \pacm\ because our bound is not on the evidence marginal $p(\{x_i\}_i^n) = \E_{r(\Theta)}[ \prod_i^n p(x_i|\Theta)]$ but rather on the \emph{posterior predictive distribution}, $p(X|\{x_i\}_i^n)=\E_{q(\Theta|\{x_i\}_i^n)}[p(X|\Theta)]$.  
In fact, in the limit $m \to \infty$ we already explicitly encode the idea that the actual Bayesian posterior is not directly useful.


\pacm\ offers real benefits in predictive performance 
in cases of model misspecification, in particular,
when a mixture of our model family would be a better predictive model 
than the model itself.
If so, why not simply fit mixture models?
This certainly does work, as
\cref{fig:masegosa_mixture_wellsp} demonstrates. 
Mixtures have proven difficult to fit
in general~\citep{morningstar2020automatic}.
The \pacm\ family of risks subsume
classic risks and 
offer theoretical generalization guarantees
in cases of model misspecification, 
while remaining computationally tractable.
For more discussion of the differences between mixtures of various sorts, please see~\cref{app:faq}.

\section{EXPERIMENTAL RESULTS}
\label{sec:experiments}


Here we demonstrate that our new risk: \pacm, can achieve 
better predictive performance on a suite of tasks.
In all experiments, we compare the \pacm-Bayes risk ($\pacpredrisk$) to alternative objectives, including the \pac-Inferential Risk ($\pacinfrisk$), also called the Evidence Lower Bound (\elbo),
since it is a lower bound on the marginal likelihood. 
We also compare to alternative \pac-style bounds on the predictive risk, namely the \pactwot\ objective proposed in \cite{masegosa2019learning}.  



\textbf{Toy Experiments: } 
\label{sec:sinusoid}
We start with a series of three simple regression tasks designed to test three different flavors of model misspecification.  The first task is to predict data from a sinusoidal model when the variance is underestimated.  The second task is to predict data from a sinusoidal model with heteroskedastic noise, where we have assumed homoskedastic noise.  The third task is to predict data from a mixture, assuming a unimodal posterior and likelihood.  For all models we used $10^4$ draws from our data generating distribution as training points, and fit the data using a 3 layer MLP with 40 hidden units.  We used normal prior and posterior distributions over parameters, and a normal distribution with fixed variance for the likelihood. Full details of the predictive models and training procedures can be found in the supplement.  

\begin{figure*}[!htb]
    \centering
    \includegraphics[width=\textwidth]{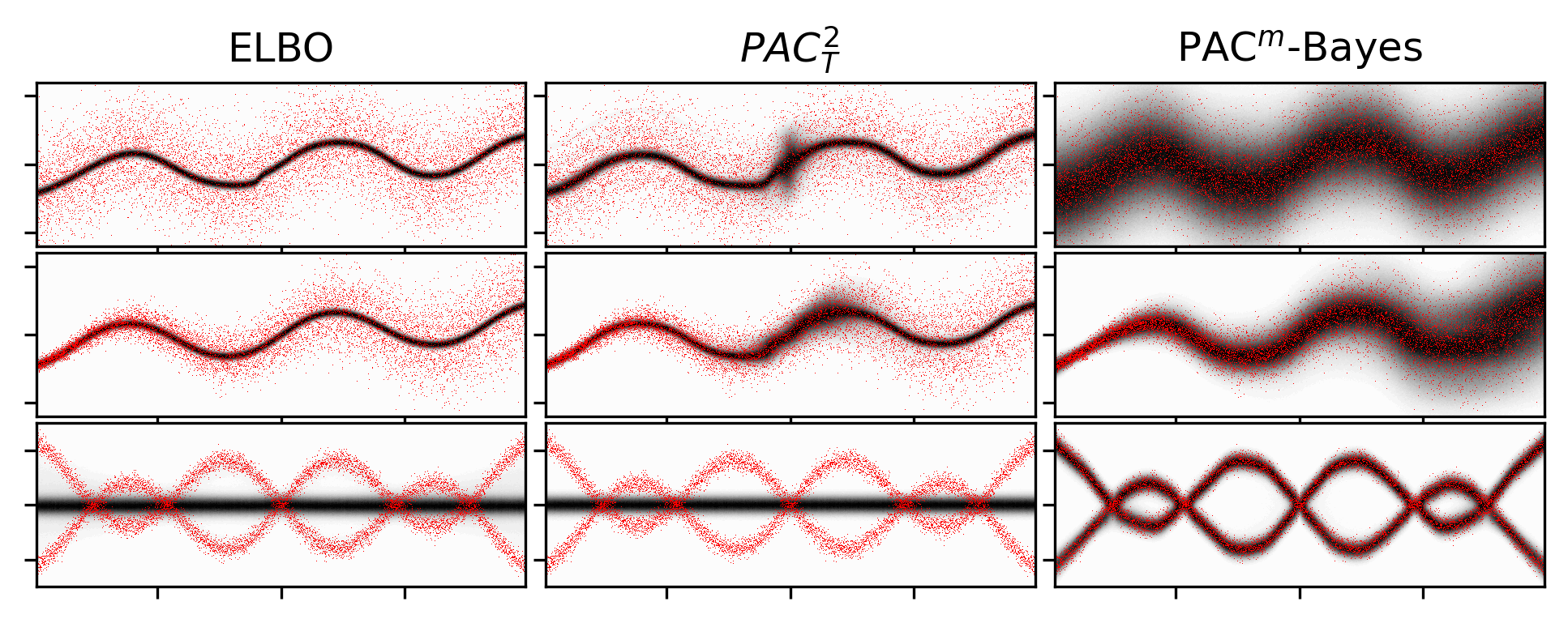}
    \caption{\footnotesize A comparison between \elbo, \pactwot, and \pacm\-Bayes on three different toy regression problems where the model is misspecified.  Datapoints are shown in red, and the black contours show the posterior predictive distribution learned by training with the loss indicated at the top of the corresponding column.  Each row covers a different data generation process and each column covers a different training loss.  All experiments use the same model family, computation, and optimization procedure, and only differ in their loss.  Despite this, \pacm\ correctly approximates the predictive distribution, while \elbo\ and \pactwot\ significantly underpredict the variance of the data in the first two cases, and do not correctly predict the modes of the output in the last case.
    \label{fig:toy_experiments}
    \label{fig:masegosa_mixture_1comp}
    }
\end{figure*}

We show the predictive models along with the data in Figure~\ref{fig:toy_experiments}.  Our first observation is that in all three cases, optimizing the \elbo\ objective leads to a poor predictive model.  This is expected behavior. Here, because the number of training points is fairly large, \elbo\ causes the posterior predictive distribution to concentrate on the (true) mean of $p(y|x)$.  Because the variance is underestimated, errors in the prediction of the mean are penalized more aggressively, exacerbating this concentration.  In other words, incorrect specification of the model leads to increasingly overconfident (but wrong) predictions when training with \elbo.  The worst-case scenario of this can be seen in the mixture experiments, when the mean of the data is often not a reasonable prediction of any of the data.

Our second observation is that while \pactwot\ does a marginally better job of accounting for the observed uncertainty (its predictive distribution is slightly wider than \elbo), it still underpredicts the variance of the data.  We appear to observe that this loss results in the model expanding the tails of its predictive distribution to account for the observed variance.  At the same time, it is clear that \pactwot\ still tends to concentrate its predictions on the mean of the data.

\begin{table}[htbp]
    \centering
    \begin{tabular}{l|ccc}
        Experiment &  \elbo & \pactwot & \pacm \\ \hline
        Sinusoid & 46.2 & 2.2 & \textbf{0.2} \\
        Heteroskedastic & 20.33 & 1.16 & \textbf{0.03} \\
        Mixture & 14.21 & 11.57 & \textbf{0.15}
    \end{tabular}
    \caption{{\footnotesize KL divergences between the learned posterior predictive model and the true predictive model.  These were each computed using 1000 samples from the learned surrogate posterior distribution.}}
    \label{tab:toy_kls}
\end{table}

In all cases, we observe that \pacm\ results in a better predictive distribution than the alternatives.  We quantitatively assess the performance using the KL-Divergence between the posterior predictive distribution, and the true generative distribution.  These are presented in Table~\ref{tab:toy_kls}.  In all cases, \elbo\ performs the worst, while \pactwot\ performs better and \pacm\ performs best.  Interestingly, we see in Figure~\ref{fig:toy_experiments} that \pacm\ recovers the multimodal posterior predictive distribution, despite the model having a unimodal posterior, prior, and likelihood.  This indicates that the surrogate posterior learned by \pacm\ is in some sense richer than that learned by \elbo, since it appears to exploit the architecture of the network in order to use its unimodal posterior to model multimodal data.  To assess if this improvement can be simply replicated by using a more expressive posterior, we repeated the mixture experiment using a Mixture of Independent Normal distributions for the posterior in~\cref{fig:masegosa_mixture_2comp}.  Here also, we find that \pacm\ results in a better approximation to the true predictive model, having a lower KL divergence from the generative distribution ($\KL=0.63$) than the alternatives ($\KL=14.19$ for \elbo\ and $\KL=9.09$ for \pactwot).  We also verify that all models perform well when the model is well specified in~\cref{fig:masegosa_mixture_wellsp}, which for the mixture problem would mean having a two component likelihood.

 \begin{figure*}[htbp]
     \centering
     \includegraphics[width=\textwidth]{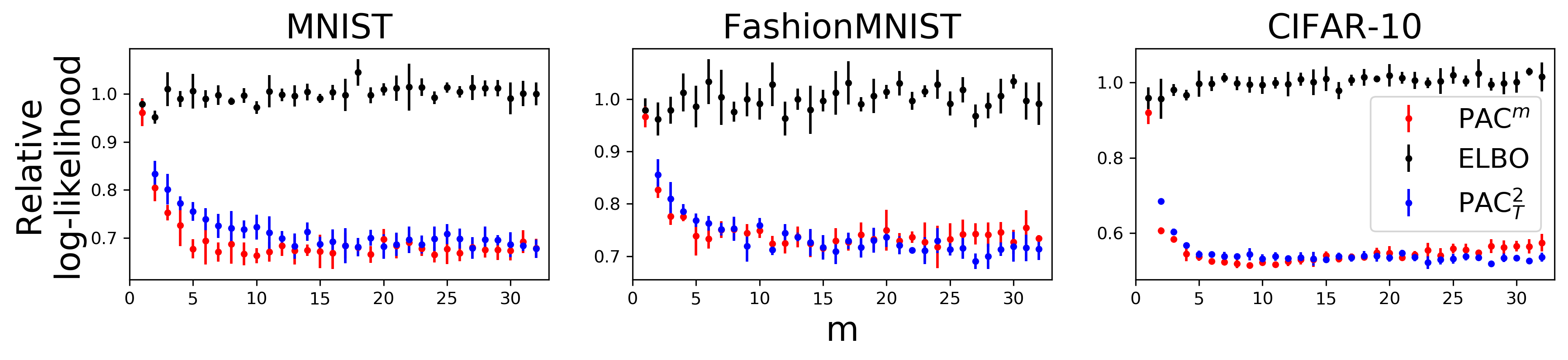}
     \caption{\footnotesize Test set log-posterior-predictive, relative to the mean log-posterior-predictive recovered using \elbo.  Black points show models trained using \elbo, while red points show models trained using \pacm\ and blue points show models trained with \pactwot.  The x-axis shows the number of samples used in computing the loss ($m$).  We see that \pacm\ appears to offer the best log-posterior-predictive, but that performance either saturates or degrades as the sample number becomes to large, indicating issues with the optimization procedure.}
     \label{fig:structured_prediction}
 \end{figure*}

\textbf{Structured Prediction: }
We also test our objective on structured prediction tasks \citep[e.g.][]{sohn2015learning}.  For this, we train a Bayesian neural network to predict the bottom half of an image, using only the top half as an input.  We test this on 3 different image datasets:  MNIST \citep{lecun1998mnist}, FashionMNIST \citep{xiao2017fashion}, and CIFAR-10 \citep{krizhevsky2009learning}.  For our likelihood, we use a Normal distribution where each pixel is considered independent.  Following \cite{masegosa2019learning}, we further fix the scale of the likelihood distribution to 1/255.  These choices are interesting for two reasons.  First, this setup also replicates the training setup which is often employed in training naive Variational Autoencoders \citep[see e.g., ][]{kingma2013auto, tomczak2018vae}, where the output variance is either fixed or shared between pixels (for non binarized images) and where all output pixels are assumed to be independent.  Second, this setup is a misspecified model since we know that the pixels in the data are not independent, at least not at the granularity which we are able to capture in most models.  We therefore hypothesize that \pacm\ should be able to offer improvements in predictive performance.

We train models and measure performance as a function of $m$ and the loss function used in training.  For each value of $m$ and each loss, we conduct 5 trials with different initializations to estimate the uncertainty in our final test set negative log-posterior-predictive probability.  We show the results in Figure~\ref{fig:structured_prediction}.  We find that the performance of \elbo\ is roughly static in $m$, with much of the observed variation consistent with noise.  This is consistent with our expectation.  In contrast, \pactwot\ and \pacm\ exhibit rapid improvement in performance with $m$, showing that the model is, in fact, misspecified and that these models are therefore able to offer meaningful improvement in predictive performance.  Interestingly, we also appear to observe a saturation in $m$ when the number of samples approaches the size of the batch.  This could occur for two reasons.  First, as we show in \cref{thm:psi_bound}, the model may ultimately cease to improve in $m$ because the increasing value of $\psi$ may overcome the tightening of $\pacpredrisk$.  Alternatively, this could be due to empirical variance in the gradients introduced by minibatch training.  This is similar to the findings from \cite{rainforth2018tighter} who showed that variance in the gradients results in an impedance to effective learning which eventually overcomes tightness.  Alternative gradient estimators such as that from \cite{tucker2018doubly} may help to solve this issue.


\textbf{Classification: }
So far, we have experimented with models where the likelihood was  either purposefully misspecified in order to highlight the generalization gap introduced by minimizing $\pacinfrisk$ compared to \pacm\, or where we expect that it is misspecified because the assumptions we make about the output data are likely to be incorrect (e.g. pixels are likely non-independent in most images).   
It is unclear the degree to which this is an issue for many real-world applications where we use highly expressive deep neural network models, but in many cases we are still forced to make incorrect modeling assumptions for the sake of convenience.  


It is equally interesting to consider the performance of \pacm\ when the likelihood is well specified, but when other parts of the Bayesian model (the prior) are not.  A good example of such a scenario is image classification, where we expect that a categorical distribution is a reasonable choice of likelihood.
To test this scenario, in~\cref{sec:moreexperiments} we present 
additional experiments where we use Bayesian convolutional neural networks to classify images from the datasets used in the previous section. 
We consider two cases:  (1) being Bayesian over the weights of the model (the ``global'' variables), 
or (2) being Bayesian over the activations of the model (the ``local'' variables).  
This latter case has been explored in works such as \cite{alemi2016deep}. 
Here we consider the same approach, except where we minimize a \pac-Bayesian bound on the predictive likelihood.


As we expect, for classification problems, we find that though the model appears to be mostly well-specified, \pacm\ learns models that make better predictions at the same cost, measured in terms of the $\KL$ divergence between the posterior and the prior.

\section{CONCLUSION}

Something as simple as a one line change to a variational Bayes setup can have drastic effects.
Swapping an expected log likelihood across multiple draws from a variational posterior with the log of the expected likelihood can vastly improve the predictive performance of badly misspecified models (summarized in~\cref{app:quickref}).  In this work we attempted to explain this phenomenon. 

Bayesian inference minimizes a stochastic upper bound on the predictive risk but the tightness of this bound is limited by model misspecification.
In this work we proposed \pacm, a new bound that directly targets predictive risk.
We demonstrated that \pacm\ outperforms \elbo\ and \pactwot\ \citep{masegosa2019learning} on misspecified Bayesian models on a wide set of example problems.


\subsubsection*{Acknowledgements}
Acknowledgements.
We would like to thank Ben Poole, Sergey Ioffe, and Rif A Saurous for useful comments.  We would also like to thank the reviewers of both this and previous versions of the paper, whose constructive feedback helped us make substantive improvements to this work.

\bibliographystyle{plainnat}
\bibliography{references}

\clearpage
\appendix
\thispagestyle{empty}
\pagebreak[2]
\makeatletter
\@addtoreset{theorem}{section}
\makeatother
\onecolumn
\makesupplementtitle

\section{Quick Reference: Comparing Different Losses}
\label{sec:different_losses}
\label{app:quickref}

Here we depict several losses closely related to \pacm\ and highlight their structural similarities and differences. The likelihood is $p(y|Z)$ and the prior/posterior discrepancy term is $\frac{r(Z)}{q(Z)}$. 

\definecolor{firebrick}{HTML}{B22222}
\definecolor{steelblue}{HTML}{4682b4}

\definecolor{col1}{HTML}{357933}  
\definecolor{col2}{HTML}{984EA3}
\definecolor{col4}{HTML}{E41A1C}
\definecolor{col3}{HTML}{377EB8}

\definecolor{col3}{HTML}{000000}
\definecolor{col4}{HTML}{000000}

\begin{align*}
\textsf{ELBO} &\defeq -\E_{q(Z^m)}\left[
  \avglog{\textcolor{col3}{p(y|Z_j)} 
  }
  + \avglog{\textcolor{col4}{\frac{r(Z_j)}{q(Z_j)}}}
  \right]\\
\text{\pacm} &\defeq -\E_{q(Z^m)}\left[
  \logavg{\textcolor{col3}{p(y|Z_j)}}
  + \frac{1}{\beta}\avglog{\textcolor{col4}{\frac{r(Z_j)}{q(Z_j)}}}
  \right] \\
\textsf{IWAE} &\defeq -\E_{q(Z^m)}\left[
  \logavg{ \textcolor{col3}{p(y|Z_j)} \textcolor{col4}{\frac{r(Z_j)}{q(Z_j)}}} \right]\\
\text{\pactwot} &\defeq \textsf{ELBO} - \E_{q(Z^m)}\left[ \textsf{SampleVariance}(y, Z^m) \right] 
\end{align*}

Where we colored the \textcolor{col1}{average-log}, \textcolor{col2}{log-average}
terms,
re-framed all losses in terms of multiple samples, and where:
\begin{itemize}
\item \textsf{ELBO} is the evidence lower bound (as a loss). \citep{blei2017variational}
\item \textsf{IWAE} is the importance weighted autoencoder loss of \citet{burda2015importance}.
\item \pactwot\ is the loss of \citet{masegosa2019learning}.
\end{itemize}
\section{FAQ}
\label{app:faq}

Below we informally address some readers' questions.

\subsection{Why did you choose the names \emph{predictive} and \emph{inferential} risk?}

These terms serve to distinguish between making the best prediction vs choosing the best model parameters.  We chose the name \emph{predictive risk} risk because this term corresponds with (among other things) the posterior \emph{predictive} distribution.  The term \emph{inferential risk} was chosen to reflect a judgement of the likelihood of model parameters.  While we realize that the machine learning community sometimes uses \emph{inferential risk} in contexts which we call \emph{predictive risk}, we felt that our use of the term has precedent in statistics and is therefore not unreasonable. 

\subsection{Is the bound non-vacuous?}
\label{sec:vacuous}

For any fixed $m$: no. When $m$ is fixed we have a slack term in our PAC bound similar to that seen in most other PAC bounds. As we show in~\cref{app:pacmproof}, the bound slack can grow at best like $o(\log m)$ or--as we've opted in the paper--like $o(m)$.  (If the bound grew in some polynomial of $n$ this would indeed be a worrisome, if not vacuous result.)

The reason it is ``ok'' that the bound slack grows in $m$ is because taking the limit of $m \to \infty$ is not an interesting nor recommended limit. Recall that the Bayesian formalism stipulates that $m=1$ and that if the model is well-specified, the Bayesian posterior is \emph{optimal}.  Choosing $m>1$ serves only as a stopgap to improve--but not cure--poor predictive risk guarantees due to the misspecified regime.

We believe that ``relaxing'' the Bayesian posterior in the sense of $m>1$ is a valuable contribution to the ML and statistics communities because it is a small step toward bridging the impressive predictive performance of the frequentist methodology with the impressive explainability/transparency of the Bayesian methodology.

\subsection{Can we recover the correct parameter value, if one exists?}

This question--while very interesting--is intentionally out of scope for this research. One technique to answer this question would be to explore the asymptotic consistency/efficiency similar to the LeCam style proofs of maximum likelihood.  However this would be a very different set of tools than the PAC approach and carry its own set of worries. I.e., ``is asymptotic analysis 'reasonable?''' and ``are typical regularity assumptions justified in the deep learning era?''

\subsection[How is PACm different from using a mixture posterior?]{How is \pacm\ different from using a mixture posterior?}

The \pacm\ loss could work \emph{on top} of a mixture posterior (or not).  That is, \pacm\ simply repeatedly samples from whatever posterior family is being fit.
We illustrate this in \cref{fig:masegosa_mixture_2comp}.

In general one should expect better performance from 
either \elbo\ or \pacm\ when using a richer posterior family.  However, as our experiments and analysis shows, \elbo\ predictive risk is more sensitive to misspecification in the likelihood and/or prior.

\subsection[How is PACm different from using a mixture likelihood?]{How is \pacm\ different from using a mixture likelihood?}

The \pacm\ loss is used to identify a distribution over the likelihood parameters, be the a likelihood mixture distribution or otherwise.  In this sense, using a mixture likelihood and using \pacm\ are largely orthogonal changes to the modeling setup, though they can be done to accomplish the same objectives.

That said, \pacm\ seems to be particularly helpful when the likelihood and/or posterior family lacks the capacity to capture multimodality present in the data.  Compare the results in \cref{fig:masegosa_mixture_1comp} to the results in \cref{sec:wellspecified}.
Taking the same misspecified model and switching to the \pacm\ objective is enough to fit multimodal 
data well.  
Fitting a proper mixture model (as in modifying the likelihood to have two components) to the multimodal data works correctly as shown in \cref{sec:wellspecified}.  In this case the model is well specified.  Details for this experiment can be found in \cref{sec:mixturedetails}.  Note that when switching from the misspecified unimodal model to the well specified mixture we have increased the number of parameters in our model, additional parameters for our variational posterior to approximate. For a simple problem like this one, having our variational posterior predict two means instead of one isn't much of an additional cost, but for larger Bayesian models like Bayesian neural networks the additional burden of fitting the mixture is hard to ignore.

Overall, switching from a model to a mixture model is changing the structure of the model.  Switching from \elbo\ to \pacm\ is simply changing the loss. That simply changing the loss of a misspecified model can recover a lot of the predictive benefits of the much larger mixture model is the primary benefit \pacm\ brings.


To further highlight the distinction, consider the choices you would have to make to a Bayesian mixture in order to recover the \pacm\ objective. Starting with a mixture likelihood ($m$ component likelihoods $p_j$ with mixing weights $w_j$), and the \elbo\ objective:
\[ -\E_{q(Z)} \left[ \log \left( \sum_j^m w_j p_j(y|Z) \right)   + \frac 1 \beta \log \frac{r(Z)}{q(Z)} \right], \]
we will have to structure our mixture so that the parameters are non-overlapping.  Let $Z^m = [Z_1, Z_2, \dots]$ denote the partition of all the component's parameters and choose a prior that factorizes in the same way:
\[ -\E_{q(Z^m)} \left[ \log \left( \sum_j^m w_j p_j(y|Z_j) \right)   + \frac 1 \beta \sum_j^m \log \frac{r(Z_j)}{q(Z_j)} \right]. \]
One would now have to fix the mixture probabilities to be uniform ($w_j = \frac 1 m$) and ensure that each of the mixture components were replications of same model $(p_j = p)$:
\[ -\E_{q(Z^m)} \left[ \log \left( \frac 1 m \sum_j^m p(y|Z_j) \right)   + \frac 1 \beta \sum_j^m \log \frac{r(Z^m)}{q(Z^m)} \right]. \]
Further, not only would the distributional form of the variational approximation for each of
the components have to be the same,
but the parameters would have to be shared, i.e.
the exact same variational posterior would be used for the parameters
of each component of the mixture.
This would generate the same objective as \pacm\ (after rescaling $\beta$) at the cost of severe and specific choices.  Instead, as we demonstrated, starting with a single mismatched model and simply targeting the predictive rather and inferential risks is a better way to arrive at \pacm\ and a principled way to fit misspecified Bayesian models.

\subsection[How is PACm different than training an ensemble?]{How is \pacm\ different than training an ensemble?}

Ordinarily when people train an ensemble they minimize the Empirical Inferential Risk ($\empinfrisk$) multiple times independently, and then average the predictions of the resulting point estimates.  By targeting an inferential risk, this won't address misspecification in the way \pacm\ can.  One could target the Empirical Predictive Risk ($\emppredrisk$) directly, this is known as a non-parameteric mixture.  For certain models 
this can perform quite well, but for rich enough model families this can severely overfit. \pacm\ adds the KL regularization term that can prevent overfitting.

\subsection[How is PACm different from IWAE?]{How is \pacm\ different than IWAE?}

IWAE gives a bound on the marginal likelihood, not the predictive distribution. Summarized in
\cref{app:quickref} practically the difference is in how the ratio of the prior and posterior densities contributes.  IWAE attempts to bound the marginal likelihood, in other words the
prior predictive likelihood.  \pacm\ is a bound on the \emph{posterior} predictive likelihood.
\section{Proofs}
\label{app:proofs}
This section proves our main theoretical result (\cref{thm:pacm}) as well as presents additional theory relevant to \pacm.

\subsection{Relationship Between Predictive and Inferential Risks In the Presence of Model Misspecification}
\label{sec:wellspecifiedtight}

The following two results are adapted from \cite{masegosa2019learning} to our notation and given here for the reader's convenience. These results examine conditions under which solutions to the inferential risk, $\min_{q(\Theta)} \trueinfrisk[q],$ are equivalent to solutions to the predictive risk, $\min_{q(\Theta)} \truepredrisk[q].$
That is, these lemmas show that model misspecification introduces a gap between \emph{predictive risk} ($\truepredrisk$) and \emph{inferential risk} ($\trueinfrisk$). This gap is potentially problematic because machine learning practitioners care about $\truepredrisk$ but minimize (an approximation of) $\trueinfrisk.$

\begin{lemma}\label{lma:tight}
$\argmin_{q(\Theta)} \trueinfrisk[q] \equiv \argmin_{q(\Theta)} \truepredrisk[q]$ only if for any distribution $\rho$ over $\Theta$,  
\[ \KL[\nu(X); p(X|\theta^{(\textsf{ml})})] \le \KL[\nu(X); \E_{\rho(\Theta)}[p(X|\Theta)]],\]
and $q^{(\textsf{ml})}(\Theta) = \argmin_{q(\Theta)} \trueinfrisk[q] \equiv \delta(\Theta - \theta^{(\textsf{ml})})$ where $\delta$ is the Dirac-delta distribution.  
\begin{proof} \emph{(Sketch.)}
Note that,
\begin{align*}
\truepredrisk[q] 
&= \KL[\nu(X), \E_{q(\Theta)} p(X|\Theta)] + \H[\nu(X)] \\
&\le \E_{q(\Theta)} \KL[\nu(X), p(X|\Theta)] + \H[\nu(X)] \\
&= \trueinfrisk[q]
\end{align*}
where the inequality is Jensen's.
Since the theorem condition implies $\trueinfrisk[q^*] \le \truepredrisk[q]$ then $\trueinfrisk[q^*] \le \min_q \truepredrisk[q] \le \trueinfrisk[q^*]$ and the claim follows.
(See Lemma~2 of \cite{masegosa2019learning} for original proof; our sketch is based on a sandwich argument.)
\end{proof}
\end{lemma}

\begin{lemma}\label{lma:loose}
If there exists a density $\rho$ over $\Theta$ such that 
\[ \KL[\nu(X); \E_{\rho(\Theta)}[p(X|\Theta)] < \KL[\nu(X); p(X|\theta^{(\textsf{ml})}], 
\]
then a minimizer of $\trueinfrisk$  is not a minimizer of $\truepredrisk$
where 
\[ q^{(\textsf{ml})}(\Theta) \defeq \argmin_{q(\Theta)} \trueinfrisk[q] \equiv \delta(\Theta - \theta^{(\textsf{ml})}), \]
where $\delta$ is the Dirac-delta distribution.
\begin{proof} \emph{(Sketch.)}
The condition of this lemma implies that $q^{(\textsf{ml})}(\Theta)$ cannot be a minimizer of $\truepredrisk$ however it is the minimizer of $\trueinfrisk.$
(See Lemma~3 of \cite{masegosa2019learning} for original proof.)
\end{proof}
\end{lemma}

\subsection[PAC-Bayes]{\pac-Bayes Relationships}
\label{sec:pacbayesproof}

This section presents two well-known \pac-Bayes results as special cases of \cref{thm:pacm}.

\begin{corollary} Under the conditions of \cref{thm:pacm}, then with probability at least $1-\xi$, $\truepredrisk[q] \le \pacinfrisk_n[q; r, \beta] + \psi_1.$
\begin{proof}
Immediate from \cref{thm:pacm} when $m=1$.
\end{proof}
\end{corollary}

\begin{corollary}
The Bayesian posterior $p(\Theta|\{x_i\}_i^n) \propto r(\Theta) \prod_i^n p(x_i|\Theta)$ minimizes \pacm\  when $m=\beta=1.$
\begin{proof}
\pacm\ is equivalently \pac\ when $m=\beta=1$ for which the claim is proven by \cite{germain2016pac}.
\end{proof}
\end{corollary}

\subsection[PACm]{\pacm-Bayes Theory}
\setcounter{theorem}{0}

\label{app:pacmproof} 


\begin{theorem}
\label{thm:pacmproof} 
    For all $q(\Theta)$ absolutely continuous with respect to $r(\Theta)$, $X^n\iid \nu(X)$, $\beta \in (0,\infty)$, $n,m\in \mathbb{N}$, $p(x|\theta) \in (0,\infty)$ for all $\{x \in \mathcal{X}:\nu(x)>0\} \times \{\theta \in \mathcal{T}: r(\theta)>0\}$, and $\xi \in (0,1),$ then with probability at least $1-\xi,$
    \begin{equation}
        \truepredrisk[q] \leq \pacpredrisk_{m,n}[q;r,\beta]  
        + \psi(\nu, \beta, m, n, r, \xi)
        - \tfrac{1}{\beta m n} \log \xi
    \end{equation}
    and furthermore (unconditionally),
    \begin{equation} 
        \pacpredrisk_{m,n}[q;r,\beta] 
        \leq \pacpredrisk_{m-1,n}[q;r,\beta]
        \leq \pacpredrisk_{1,n}[q;r,\beta]
        =   \pacinfrisk_n[q,r,\beta] 
         \tag{\ref{eqn:thm1-2}}
    \end{equation}
    where:
    \begin{align}
    \pacpredrisk_{m,n}[q;r,\beta] 
      &\defeq
       - \frac{1}{n} \sum_i^n \E_{q(\Theta^m)}\left[  \log \left( \frac{1}{m} \sum_j^m p(x_i | \Theta_j) \right) \right] + \frac{1}{\beta n} \kl{q(\Theta)}{r(\Theta)}
       \defeq \textrm{\pacm}  \tag{\ref{eqn:pacpredrisk}} \\
    \psi(\nu, \beta, m, n, r, \xi) &\defeq
      \tfrac{1}{\beta m n} \log \E_{\nu(X^n)}  \E_{r(\Theta^m)} \left[ e^{\beta n m \Delta(X^n, \Theta^m)}  \right]  \tag{\ref{eqn:psi}} \\
    \begin{split}
        \Delta(X^n, \Theta^m) 
        &\defeq
      \frac{1}{n}\sum_i^n \log \left( \frac{1}{m} \sum_j^m p(X_i|\Theta_j) \right) -\E_{\nu(X)}\left[\log \left( \frac{1}{m} \sum_j^m  p(X|\Theta_j) \right) \right].
     \end{split}
       \tag{\ref{eqn:gap}}
    \end{align}
\begin{proof}



Write:
\begin{align*}
g(\Theta^m; X) &\defeq \frac{1}{m}\sum_j^m p(X | \Theta_j) \\ 
\overline{\joshdevice}_{n,m}[q] &\defeq -\frac{1}{n} \sum_i^n \mathbb{E}_{q(\Theta^m)}\left[  \log g(\Theta^m; x_i) \right]  \\
\joshdevice_m[q] &\defeq -\E_{\nu(X)}\mathbb{E}_{q(\Theta^m)}\left[  \log g(\Theta^m; X) \right]
\end{align*}


For the first claim:

\begin{quote}
Jensen's inequality implies \[
-\log \E_{q(\Theta^m)}\left[ g(\Theta^m; X) \right] \le \E_{q(\Theta^m)}\left[ -\log g(\Theta^m; X) \right]. 
\]
Applying $\E_{\nu(X)}$ to both sides
implies $\truepredrisk[q]  \le \joshdevice_m[q].$

To complete the proof of the first claim, we now show 
\[ p(\joshdevice_m[q] \le \pacpredrisk_{n,m}[q;r,\beta] + \psi_{n,m}) \ge 1 - \xi.
\]
Make the substitution, $f(\Theta^m;\{x_i\}_i^n) \defeq \beta m n \Delta(\{x_i\}_i^n, \Theta^m)$ (for some non-stochastic $\beta m n$) to Lemma~\ref{lma:compression} (``Compression Lemma'') and rearrange:
\begin{align*}
- \E_{q(\Theta^m)} & \E_{\nu(X)} [\log g(\Theta^m;X)]
  \le -\E_{q(\Theta^m)} \E_{\nu(X|\{x_i\}_i^n)} [\log g(\Theta^m;X)] \nonumber \\
 &\quad
     + \tfrac{1}{\beta m n} \KL[q(\Theta^m), r(\Theta^m)]
     + \tfrac{1}{\beta m n} \log \E_{r(\Theta^m)} \left[ e^{\beta m n\Delta(\{x_i\}_i^n, \Theta^m)} \right].
\end{align*}

For the $\KL$ term, note that Lemma~\ref{lma:kliid} (``$\KL$-divergence iid'') implies 
\[ \KL[q(\Theta^m), r(\Theta^m)] = m \KL[q(\Theta), r(\Theta)]. \]

For the rightmost term (a log moment generating function conditioned on $\{x_i\}_i^n$), make substitutions $Z \defeq \E_{r(\Theta^m)} [ e^{\beta m n\Delta(\{x_i\}_i^n, \Theta^m)}]$ and $p \defeq \nu(X^n)$ to Lemma~\ref{lma:logmarkov} (``Log Markov Inequality'') to conclude:
\begin{multline*}
\nu_{\text{\scalebox{0.65}{$X^n$}}}\left(\log \E_{r(\Theta^m)} \left[ e^{\beta m n\Delta(X^n, \Theta^m)} \middle| X^n \right] 
\right.
\left. \le
\log \E_{\nu(X^n)} \E_{r(\Theta^m)} \left[ e^{\beta m n\Delta(X^n, \Theta^m)} \right] - \log \xi \right)  \ge 1-\xi.
\end{multline*}

Scale the inner inequality by $\frac{1}{\beta m n}$ (which doesn't change the probability) and combine this result with the previous two to prove the first claim.

(This proof was inspired by \cite{masegosa2019learning}.)
\end{quote}

For the second claim:

\begin{quote}

Note that the $\KL$ terms of $\pacpredrisk_{n,m}$ and $\pacinfrisk_n$ are not functions of $m$ and can be ignored. The equality $\pacpredrisk_{n,1}[q;r,\beta] = \pacinfrisk_n[q,r,\beta]$
is true by definition; $g(\Theta^1; X) = p(X|\Theta).$ To complete the proof it is sufficient to show $\overline{\joshdevice}_{n,m} \le \overline{\joshdevice}_{n,m-1}.$ I.e.,
\begin{align*}
\overline{\joshdevice}_{n,m}[q]
&= -\frac{1}{n}\sum_i^n \E_{q(\Theta^{m})} \left[ \log \frac{1}{m} \sum_j^{m} p(x_i|\Theta_j) \right]\\
&= -\frac{1}{n}\sum_i^n \E_{q(\Theta^{m})} \left[ \log \frac{1}{m} \sum_j^{m} \frac{1}{m-1}\sum_{k\ne j}^{m}p(x_i|\Theta_k) \right]\\
&\le -\frac{1}{m} \sum_j^{m} \frac{1}{n}\sum_i^n  \E_{q(\Theta^{m})} \left[ \log  \frac{1}{m-1}\sum_{k\ne j}^{m}p(x_i|\Theta_k) \right]\\
&= \frac{1}{m} \sum_j^{m} \overline{\joshdevice}_{n,m-1}[q] \\
&= \overline{\joshdevice}_{n,m-1}[q].
\end{align*}
The inequality is Jensen's and the second-to-last equality follows from $\Theta^{m}$ being independent.

(This proof is inspired by \cite{burda2015importance}.)
\end{quote}
\end{proof}
\end{theorem}

While \Cref{thm:pacmproof} is technically true, additional assumptions are needed to ensure it is nonvacuous, i.e., $\psi(\nu, \beta, m , n, r, \xi) < \infty.$ \Cref{thm:psi_bound} (below) affirms this is the case when $\Delta(X, \theta)$ is \emph{everywhere $s^2$-sub-gaussian} for all $\{\theta \in \mathcal{T} : r(\theta)>0\}$ and furthermore suggests that:
\begin{itemize}
\item $\beta_n=O(1)$ implies $\psi_n=O(1)$
\item $\beta_{m,n}=O(1)$ implies:
  \begin{align}
    \psi_{m,n}&=O(m) && \text{(Larger slack.)}\\
    (\beta n)^{-1}\mathsf{KL}&=O(n^{-1}) &&\text{(Weaker $q$ regularization.)}\\
    (\beta n m)^{-1}\log \xi &= O(m^{-1}n^{-1}) && \text{(Faster error decay.)}
  \end{align}
\item $\beta_{m,n}=O(m^{-1})$ implies:
  \begin{align}
    \psi_{m,n}&=O(\log m) && \text{(Smaller slack.)}\\
    (\beta n)^{-1}\mathsf{KL}&=O(m n^{-1}) && \text{(Stronger $q$ regularization.)}\\
    (\beta n m)^{-1}\log \xi &= O(n^{-1}) &&\text{(Slower error decay.)}
  \end{align}
\end{itemize}

We emphasize that  sub-gaussianity is only assumed for $n=m=1,$ yet our proof holds for $n,m\ge 1.$ This assumption is similar to that made by \citet{germain2016pac}, however we assume $\Delta(X,\theta)$ is everywhere sub-gaussian whereas they assume $\Delta(X,\Theta)$ is \emph{jointly sub-gaussian}.  We note that their Corollaries 4 and 5 (the relevant claims) have incorrect proofs which do not obviously follow from joint sub-gaussianity; our \Cref{thm:psi_bound} with $m=1$ serves as a correction and also explains the different technical assumption. As also indicated in \citet{germain2016pac}, our $\psi$'s finiteness is also provable by stronger assumptions, e.g., $p(x|\theta) \in [a,b]$ where $a,b \in \mathbb{R}_{\ge 0}$ and for all $\{x \in \mathcal{X}:\nu(x)>0\} \times \{\theta \in \mathcal{T}: r(\theta)>0\}$. \citep{catoni2007pac,alquier2016properties}  However, we refain from making such claims, preferring the arguably more general assumptions of \Cref{thm:psi_bound}.

\begin{theorem}\label{thm:psi_bound} 
Making the assumptions of \Cref{thm:pacmproof} and additionally that for all $\{\theta \in \mathcal{T} : r(\theta)>0\},$ $\Delta(X,\theta)$ is sub-gaussian with standard deviation $0 < s_\theta \le s < \infty,$ i.e.,
$\log \E_{\nu(X)}\left[ e^{\lambda \Delta(X,\theta)} \right] \le \tfrac{1}{2}s_\theta^2\lambda^2 \le \tfrac{1}{2}s^2\lambda^2,$ then:
\begin{align}
\psi_{m,n}
&= \tfrac{1}{\beta m n} \log \E_{\nu(X^n)} \E_{r(\Theta^m)}\left[ e^{\beta m n \Delta(X^n, \Theta^m) } \right]  \\
&\le \tfrac{1}{2}s^2 \beta m  +\left(1+\frac{1}{\beta m}\right)\log m. \label{eq:psi_bound}
\end{align}

Additionally,
\begin{equation}
\beta^* = m^{-1} s^{-1} \sqrt{2\log \max(e,m)},
\end{equation}
minimizes \cref{eq:psi_bound} for $m>1$ and is a constant when $m=1$, i.e., \cref{eq:psi_bound} at $\beta^*$ is,
\begin{equation}
 \psi_{n,m}
\le 
\frac{s}{\sqrt{2}} \left(
  \left(\log \max(e,m)\right)^{\frac{1}{2}}
  +
  \left(\log \max(e,m)\right)^{-\frac{1}{2}}
  \right)+\log m=O(\log m).
\end{equation}

\begin{proof}
Begin by noting that,

\begin{align}
\Delta(x, \{\theta_j\}_j^m) &\defeq \log \frac{1}{m} \sum_j^m p(x|\theta_j) - \E_{\nu(X)}\left[\log \frac{1}{m}\sum_j^m p(X|\theta_j) \right] \\
&\le \max \Big\{ \log p(x|\theta_j) \Big\}_j^m - \E_{\nu(X)}\left[ \log \frac{1}{m}\sum_j^m p(X|\theta_j)\right] \\
&= \max \left\{ \log p(x|\theta_j) - \E_{\nu(X)}\left[ \log \frac{1}{m}\sum_k^m p(X|\theta_k)\right] \right\}_j^m  \\
&\le \max \Big\{ \log p(x|\theta_j) - \E_{\nu(X)}\left[ \log p(X|\theta_j)\right] \Big\}_j^m  + \log m \\
&= \max \Big\{ \Delta(x, \theta_j)  \Big\}_j^m  + \log m.
\end{align}
The first inequality follows from the upper bound in Lemma~\ref{lma:logavgexp_bound_simple}. The second inequality follows from the negative of the lower bound in Lemma~\ref{lma:logavgexp_bound_simple}, i.e., \[-\log \frac{1}{m}\sum_k^m p(X|\theta_k) \le -\max\{ \log p(x|\theta_j) \}_j^m + \log m \le -\log p(x|\theta_k)+ \log m, \]
for all $ k \in \{1,\ldots,m\}$.

Combining this fact and the fact that $e^{\max \{a_j\}_j^m} = \max \{e^{a_j}\}_j^m  \le \sum_j^m e^{a_j},$ implies:

\begin{align}
e^{\frac{\lambda}{n}\Delta(x, \{\theta_j\}_j^m)} 
&\le e^{\frac{\lambda}{n}\left(\max \Big\{ \Delta(x, \theta_j)  \Big\}_j^m  + \log m\right)} \\
&= m^\frac{\lambda}{n} \max \left\{ e^{\frac{\lambda}{n}\Delta(x, \theta_j)}\right\}_j^m  \\
&\le m^\frac{\lambda}{n}\sum_j^m e^{\frac{\lambda}{n}\Delta(x, \theta_j)}.
\end{align}

Combining this fact with the everywhere sub-gaussianity of $\Delta(X,\theta)$ implies:

\begin{align}
\log & \E_{\nu(X^n)}\E_{r(\Theta^m)}\left[ e^{\lambda \frac{1}{n} \sum_i^n \Delta(X_i,\Theta^m)}\right] \\
&= \log \E_{r(\Theta^m)} \left[ \prod_i^n \E_{\nu(X)} \left[ e^{ \frac{\lambda}{n} \Delta(X,\Theta^m)} \right] \right] \\
&= \log \E_{r(\Theta^m)} \left[ \left( \E_{\nu(X)} \left[ e^{ \frac{\lambda}{n} \Delta(X,\Theta^m)} \right] \right)^n \right] \\
%
%
%
&\le \log \E_{r(\Theta^m)} \left[ \left( \E_{\nu(X)} \left[ \sum_j^m e^{ \frac{\lambda}{n} \Delta(X,\Theta_j) }  \right] \right)^n \right] + \lambda \log m\\
&= \log \E_{r(\Theta^m)} \left[ \left( \sum_j^m \E_{\nu(X)} \left[  e^{ \frac{\lambda}{n} \Delta(X,\Theta_j) }  \right] \right)^n \right] + \lambda \log m\\
%
%
&\le \log \E_{r(\Theta^m)} \left[ \left( \sum_j^m   e^{ \frac{\lambda^2 s^2}{2 n^2} }  \right)^n \right] + \lambda \log m\\
&= \log \E_{r(\Theta^m)} \left[ \left( m   e^{ \frac{\lambda^2 s^2}{2 n^2} }  \right)^n \right] + \lambda \log m\\
%
%
&= \frac{\lambda^2 s^2}{2 n} + (\lambda+n) \log m
\end{align}

Scaling by $\tfrac{1}{\lambda},$ and substituting $\lambda = \beta m n$ implies $\psi_{m,n} \le
\tfrac{1}{2}s^2 \beta m  + \log m + \frac{1}{\beta m} \log m.$
Note that if $\beta=m^{-1}$ then $\psi_{m,n}\le \tfrac{1}{2}s^2 + 2 \log m.$

It now remains to find the optimal $\beta$ for $\xi=1$.
For $m\ge 3$ note that
$\tfrac{1}{2}s^2\beta m  + \tfrac{1}{\beta m}\log m+\log m$ is convex in $\beta>0$ since $m,n>0.$ Solving for the root of the gradient we find
$\beta^* = m^{-1}s^{-1}\sqrt{2\log \max(c,m)}$ where $c<3.$
For $m< 3$ we resign ourselves to finding the optimal constant above. Using $c=e$ implies that for $m<3$ then $\psi\le 2 \tfrac{1}{\sqrt{2}}s + \log(m).$
\end{proof}
\end{theorem}

\Cref{thm:psi_bound} indicates that $\beta = O(1)$ is sufficient to ensure nonvacuousness of \Cref{thm:pacm} for all $\xi,n$ and a fixed $m$ in the sense that $\psi_{n} = O(1).$ Although we emphasize that $m\to \infty$ is not a noteworthy asymptotic regime, were we to consider large $m$, then $\psi_{m,n}=O(m)$ when $\beta=O(m^{-1})$ and at best, $\psi_{m,n}=O(\log m)$ when $\beta=O(m^{-1}\sqrt{\log m}).$  That is, even for the optimial $\beta,$ $\psi$ does not vanish in $n$.  Despite these concerns we note the following:
\begin{enumerate}
\item \Cref{thm:pacm} remains non-vacuous for $\beta=O(1)$ and any finite $m$, i.e., $\psi_{m,n}$ is bounded by a constant in $s$ and $m$, analogous to the non-vanishing constant in  Corollary 4 of \cite{germain2016pac}.
\item $\psi_{m,n}$ is smallest when $\beta=O(m^{-1}\sqrt{\log(m)});$ however this choice of $\beta$ scales the $\KL$ by $m \sqrt{\log(m)^{-1}}$ which effects both accuracy and generalization (unlike changes to $\psi$ which only affects generalization).
\item In practice we recommend choosing $\beta$ by cross-validation and for each $m,n$ regime. This implies the $m,n$-parameterization is merely a theoretical consideration (especially in light of point 1 above).
\item \Cref{thm:psi_bound} is an upper bound and may or may not be made tighter. \Cref{thm:psi_bound} assumptions are arguably fairly weak and stronger assumptions might help, e.g., bounded likelihood or $\Delta(X,\{\theta\}_j^m)$ being everywhere sub-gaussian (as opposed to $\Delta(X,\theta)$ being everywhere sub-gaussian).
\item The practitioner would not typically use large $m$. Given that computational complexity also grows in $m$, we expect the vast majority of cases to use $m\le 50$ and to see improvements over $m=1$.
\end{enumerate}

\subsection{Lemmas}

In this section we present several Lemmas used to simplify this paper's proofs. Most of the Lemmas are well-known and are given here for the reader's convenience. 

\begin{lemma}[Compression]\label{lma:compression}
If $p(\Theta)$ is absolutely semicontinuous wrt $r(\Theta)$  and
$0<\E_{r(\Theta)}[e^{f(\Theta)}] < \infty$, then
$\E_{p(\Theta)}[f(\Theta)] \le \KL\left[p(\Theta), r(\Theta)\right] + \log \E_{r(\Theta)}[e^{f(\Theta)}].$
\begin{proof}
Write $q(\Theta) \defeq \frac{r(\Theta)e^{f(\Theta)}}{\E_{r(\Theta)} [
e^{f(\Theta)}]}$ and note that Lemma~\ref{lma:gibbs} implies,
$0 \le \KL\left[p(\Theta), q(\Theta)\right]
= \KL\left[p(\Theta), r(\Theta)\right] - \E_{p(\Theta)} [ f(\Theta) ] + \log \E_{r(\Theta)}[ e^{f(\Theta)} ].$
\end{proof}
\end{lemma}
Proof due to \cite{banerjee2006bayesian,zhang2006information}.

\begin{lemma}[Log Markov Inequality]\label{lma:logmarkov}
For any $\xi \in (0,1]$ and random variable $Z \sim p$ with $p(Z \le 0) = 0$ then $p(\log Z \le \log \E_p[Z] - \log \xi) \ge 1-\xi.$
\begin{proof}
Markov's inequality states that $p(Z > t) \le \frac{\E_p[Z]}{t}$ for non-negative random variable $Z \sim p$ and $t>0$. Substituting $t=\frac{\E_p[Z]}{\xi}$ implies
$p(Z > \frac{\E_p[Z]}{\xi}) \le \xi$. Combining this with the fact that
$\log$ is a non-decreasing bijection implies $p(\log Z > \log \E_p[Z] - \log
\xi)\le \xi.$ Examining the complement interval completes the proof.
\end{proof}
\end{lemma}

\begin{lemma}[$\KL$-divergence iid]\label{lma:kliid}
If $p(\Theta^m) \defeq \prod_j^m p(\Theta_j)$ and  $r(\Theta^m) \defeq \prod_j^m r(\Theta_j)$, then $\KL[p(\Theta^m), r(\Theta^m)]=m\KL[p(\Theta),r(\Theta)].$
\begin{proof}
$\KL[p(\Theta^m), r(\Theta^m)] = \E_{\prod_j^m p(\Theta_j)}\left[ \log \frac{\prod_j^m p(\Theta_j)}{\prod_j^m r(\Theta_j)}\right] = m \KL\left[p(\Theta), r(\Theta)\right].$
\end{proof}
\end{lemma}

\begin{lemma}[Gibb's Inequality]\label{lma:gibbs}
If $p(\Theta)$ is absolutely semicontinuous wrt $r(\Theta)$, then $\KL[p, q] \ge 0.$
\begin{proof}
$\KL[p,q] = -\E_{p(x)}\left[ \log \frac{q(x)}{p(x)} \right] \ge -\log \E_{p(x)}\left[\frac{q(x)}{p(x)}\right]=-\log 1=0$
where the inequality is Jensen's.
\end{proof}
\end{lemma}

\begin{lemma}[$\psi$ non-negative]\label{lma:psinonneg}
Under the conditions of \cref{thm:pacm} and if $\joshdevice_m[r],\overline{\joshdevice}_{n,m}[r] < \infty$, then $\psi(\nu, n,m,\beta,r, \xi) \ge 0.$
\begin{proof}
Jensen's inequality implies $e^{\E Z}\le \E e^Z$. Applying $\log$ to both sides (a monotonically increasing function), implies $\E Z \le \log \E e^Z$. Substitute $Z \defeq \beta n m \Delta_{n,m}$ (see \cref{eqn:gap}) and note $\E_{\nu(X^n)r(\Theta^m)}Z=0$ by definition. Finally, note $\log z \le z-1$ for $z > 0$ implies $-\log \xi \ge 0$ for $\xi \in (0,1]$.
\end{proof}
\end{lemma}

\begin{lemma}[Log-Average-Exp Bound -- Parametric] \label{lma:logavgexp_bound_parametric}
\[-\log \frac{1}{n} \sum_i^n e^{x_i} \le \begin{cases}-\tfrac{1}{\phi} \log \frac{1}{n}\sum_i^n e^{ \phi x_i} &  0 < \phi \le 1,\\-\frac{1}{n}\sum_i^n x_i & \phi = 0. \end{cases}\]
\begin{proof}
Write $\operatorname{\sf lse}(x)=\log \sum_i^n e^{x_i}$ and $\operatorname{\sf softmax}_i(a) = \exp\left(a_i - \operatorname{\sf lse}(a)\right).$

For the $\phi \in (0,1]$ case, note that $\operatorname{\sf lse}$ convexity and Jensen's inequality imply 
$\operatorname{\sf lse}\left(\phi a + (1-\phi) b\right) \le \phi \operatorname{\sf lse}(a) + (1 - \phi) \operatorname{\sf lse}(b)$ for $a,b \in \mathbb{R}^n.$
Multiplying by $-\tfrac{1}{\phi}$ and rearranging yields $-\operatorname{\sf lse}(a) \le -\frac{1}{\phi} \operatorname{\sf lse}\left(\phi a + (1-c)b\right) + \tfrac{1}{\phi}(1 - \phi) \operatorname{\sf lse}(b).$ Substituting $a=x - \log n$ and $b=-\log n$ proves the first case.

For the $\phi=0$ case, note that L'Hopital's rule implies:
\begin{align*}
\lim_{\phi \to 0} \frac{1}{\phi} \operatorname{\sf lse}\left(\phi x - \log n\right)
&= \lim_{\phi \to 0} \frac{\frac{\partial}{\partial \phi}\operatorname{\sf lse}\left(\phi x - \log n\right)}{\frac{\partial}{\partial \phi} \phi} 
 \\
 &= \lim_{\phi \to 0} \frac{\sum_i^n \operatorname{\sf softmax}_i\left(\phi x - \log n\right) x_i}{1} 
= \frac{1}{n}\sum_i^n x_i
\end{align*}
since $\lim_{\phi \to 0} \operatorname{\sf lse}(\phi x - \log n)=\lim_{\phi \to 0} \phi = 0.$ 
The bound follows from this fact, the convexity of $-\log z$, and Jensen's inequality: $-\log \frac{1}{n}\sum_i^n e^{x_i} \le -\frac{1}{n}\sum_i^n \log e^{x_i}.$
\end{proof}
\end{lemma}

Lemma~\ref{lma:logavgexp_bound_parametric} is potentially useful because it shows that minimizing $-\frac{1}{\phi}\log \sum_j^m p(X|\Theta_j)^\phi$ is still consistent with minimizing \pacm\ (i.e., $\phi=1$) in the sense that $\phi \in [0,1]$ implies an upper bound. This result might be useful for mitigating some of the gradient variance observed in the Monte Carlo approximation of \pacm\ for large $m$; this conjecture is left for future work. (We note that all experiments reported in this paper use $\phi=1$.)

Lemma~\ref{lma:logavgexp_bound_parametric} similarly exists in \cite{asadi2017alternative} though our proof differs slightly.

\begin{lemma}[Log-Average-Exp Bound -- Simple] \label{lma:logavgexp_bound_simple}

\begin{align}
\max\left(\frac{1}{n}\sum_i^n x_i, \max\{x_i\}_i^n -\log n\right) \le
\log \frac{1}{n}\sum_i^n e^{x_i}
\le \max\{x_i\}_i^n
\end{align}

\begin{proof}
For the upper bound, note that:
\begin{align*}
\log \frac{1}{n}\sum_i^n e^{x_i}
&= \max\{x_j\}_j^n + \log \frac{1}{m}\sum_i^n e^{x_i-\max\{x_j\}_j^n} \\
&\le \max\{x_j\}_j^n + \log \frac{1}{n}\sum_i^n e^{0} \\
&= \max\{x_j\}_j^n
\end{align*}

For the lower bound, note that:
\begin{equation*}
\log \frac{1}{n}\sum_i^n e^{x_i}
\ge \log e^{\max\{x_j\}_j^n} -\log n 
= \max\{x_j\}_j^n -\log n 
\end{equation*} 
and by Jensen's inequality,
\begin{equation*} 
-\log \frac{1}{n}\sum_i^n e^{x_i}
\le -\frac{1}{n}\sum_i^n \log e^{x_i}.
\end{equation*} 
\end{proof}
\end{lemma}

\section{Additional Experimental Results}
\label{sec:moreexperiments}

In this section we present additional experimental results which were omitted from the main text due to space constraints.

\subsection{Mixture}
\label{sec:mixture}

\autoref{fig:toy_experiments} shows that \pacm\ is able to accurately predict multimodal data even if the posterior, prior, and likelihood are all unimodal.  To test if this is a result of an effectively more expressive posterior induced by training with the loss, we repeated this experiment, but using an explicitly multimodal posterior distribution:  A mixture of two multivariate normals.  Details of this experiment are presented in \autoref{sec:experimentaldetails}.

\begin{figure}[htbp]
    \centering
    \includegraphics[width=\linewidth]{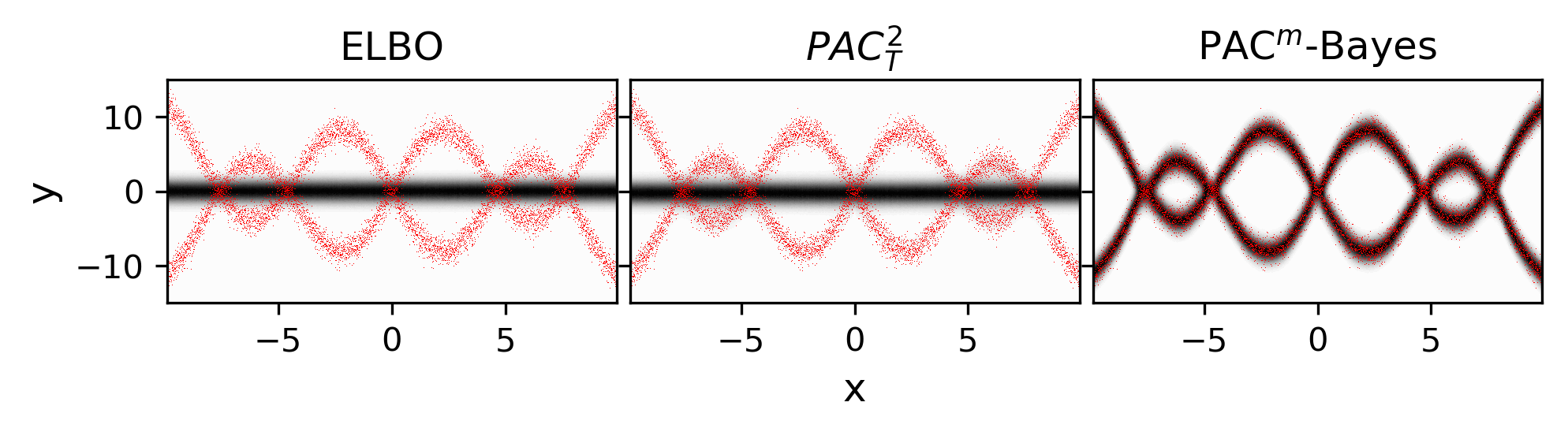}
    \caption{\footnotesize Similar to Figure~\ref{fig:toy_experiments}, but where the surrogate posterior is multimodal.  We find that the results are unchanged, despite the increased flexibility afforded to \elbo\ and \pactwot\ through the use of multiple modes in the posterior.}
    \label{fig:masegosa_mixture_2comp}
\end{figure}

In \autoref{fig:masegosa_mixture_2comp}, we show the predictive models learned by training with \elbo, \pactwot, and \pacm.  Similar to \autoref{fig:toy_experiments}, we find that \elbo\ consolidates all of its probability mass on the mean of the data, aiming to maximize the expected log-likelihood of the data (the average squared deviation between the predicted mean and observed data). \pactwot\ improves upon this slightly because the expected log-likelihood term is in tension with the variance term which tries to maximize the difference in log-likelihood between different samples of the model. 

\begin{figure}[htbp]
    \centering
    \includegraphics[width=\hsize]{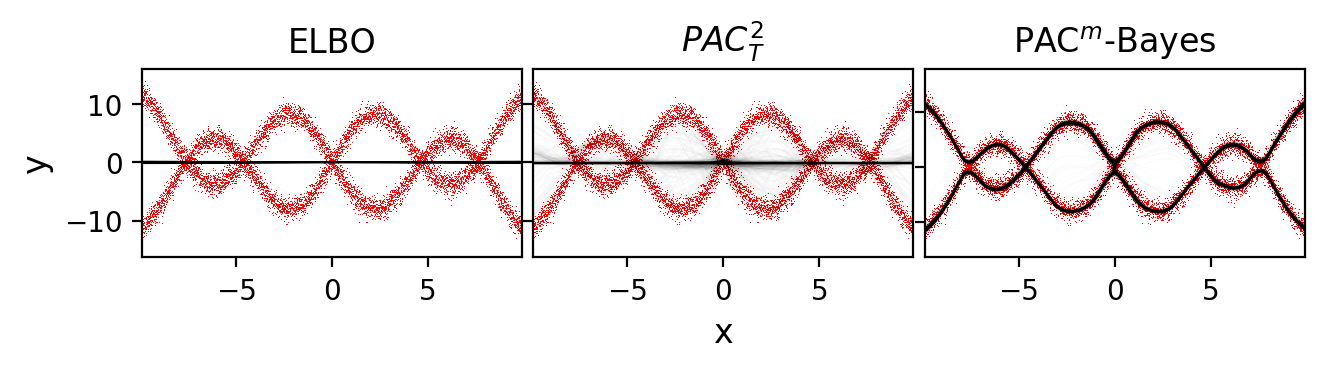}
    \includegraphics[width=\hsize]{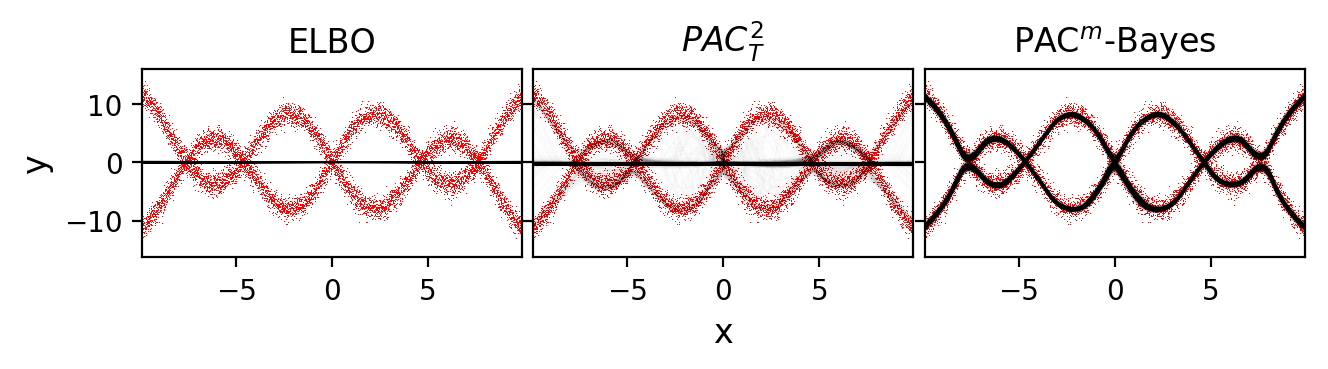}
    \caption{Visualization of the means of the likelihood predicted using samples from the posterior distribution.  The top row shows the results for a unimodal posterior, and the bottom row shows the results for a multimodal posterior.  We find that \elbo\ assigns all predictions to y=0, while \pactwot\ appears to have occasional samples that track the observed data.  \pacm\ only has samples which track the observed data.}
    \label{fig:masegosa_functions}
\end{figure}

Though we find that \pactwot\ does a much better job of predicting the data than \elbo\ (measured by the KL Divergence between the predictive model and the true generative model), this result is not obvious from looking at \autoref{fig:toy_experiments} and \autoref{fig:masegosa_mixture_2comp}.   In order to make this result more clear, instead of visualizing the histogram of samples from the predictive model, we instead draw 1000 samples from the posterior.  For each posterior sample, we show the predicted mean of the output distribution as a function of $x$.  Effectively, we want to see two curves tracing each mode in the output data.  We show the results in \autoref{fig:masegosa_functions}.  For a 1 component model, we find that \pactwot\ places most of the probability mass on the mean, with tails that can be seen reaching toward the modes.  However, we find that there are relatively few samples from the model which track any mode in the data.  For a 2 component model, we find that there are proportionally more samples which track each of the modes, but these are still much less frequent than samples which merely follow the mean in the data.  For reference, the peak probability density for samples near the data is roughly 30 times less than the density at the mean.  This is better than \elbo\ which places all of its probability near the mean.  It is also worse than \pacm, for which the 1 component model only has very few samples which fall away from either of the modes, and for which the 2 component model only has samples at each mode.

\subsection{Well Specified Mixture}
\label{sec:wellspecified}

To show that all losses perform equivalently when the loss is well specified, we show here an experiment we ran where the likelihood is assumed to be multimodal.  Here we used a mixture of normal distributions, with fixed categorical distribution and component variance.  This left us to predict only the mean, similarly to the other mixture experiments.  Additional details are presented in \autoref{sec:mixturedetails}.

\begin{figure}[htbp]
    \centering
    \includegraphics[width=\linewidth]{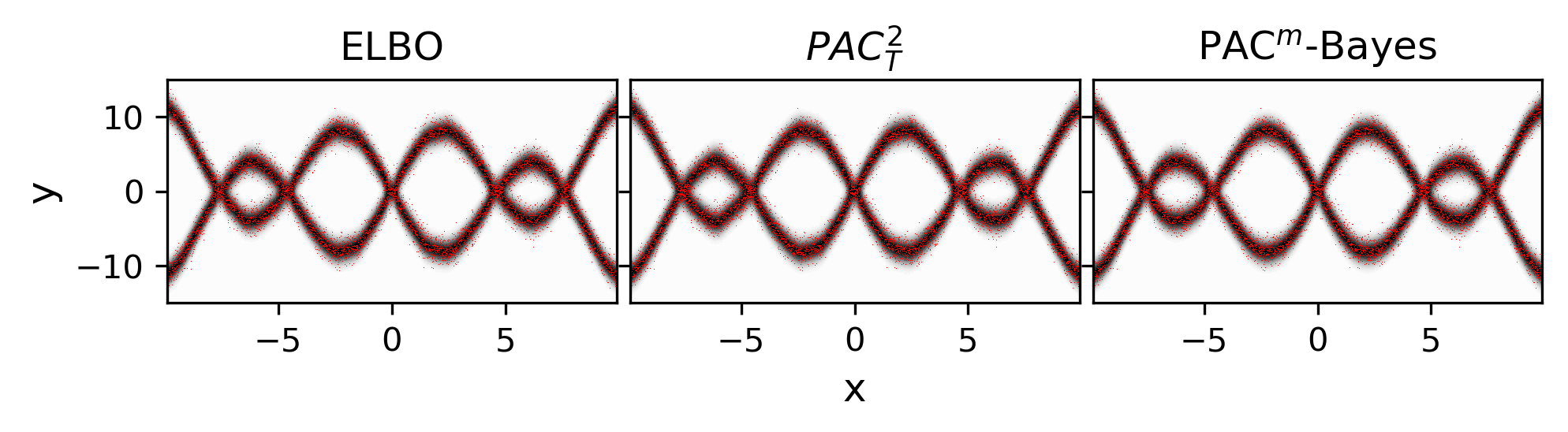}
    \caption{\footnotesize Similar to Figure~\ref{fig:toy_experiments}, but where the likelihood is multimodal.  This demonstrates performance for a well-specified model.  We see that in this scenario, all three losses recover a good predictive model.}
    \label{fig:masegosa_mixture_wellsp}
\end{figure}

We show the results in \autoref{fig:masegosa_mixture_wellsp}.  As expected, since \elbo\ is tight for well specified models, it does a reasonable job of recovering the true predictive distribution.  However, we should also note that the models which optimize bounds on the predictive risk also perform comparably.  In fact, \pacm\ observes a marginally lower KL Divergence from the true generative distribution.  We measure $\KL=0.007$ for \pacm, $\KL=0.017$ for \elbo, and $\KL=0.07$ for \pactwot.  We did not evaluate if the discrepancy is simply due to variance in the optimization or if the lower $\KL$ observed from \pacm\ is a result of optimizing a tighter bound.

\subsection{Bayesian Neural Network - Stochastic Weights}
\label{sec:weightexperiment}
For classification experiments using stochastic weights, we give the model the full images from each dataset and attempt to predict the output class.  For this, we assume the following graphical model:  
\begin{align}
&\Theta \sim r(\Theta) \\
&\text{for } i = 1 \ldots n: \nonumber\\
&\quad  y_i \sim p(Y_i | z_T(x_i, \Theta))
\end{align}
where $z_T$ is the output of a $T$-layer neural network where each layer's parameters
are specified by a partitioning of the random vector $\theta.$ For example, if $z_T$ is a multilayer perceptron, it might be defined by the recurrence $z_t(x, \theta) = a_{t-1}(z_{t-1}(x, \theta)) w_t + b_t.$ where $\{(w_t, b_t)\}_t^T$ is partition of vector $\theta$ and with appropriately reshaped members and where $a(\cdot)w$ is a (row-) vector-matrix product.

\begin{figure}[htbp]
    \centering
    \includegraphics[width=\hsize]{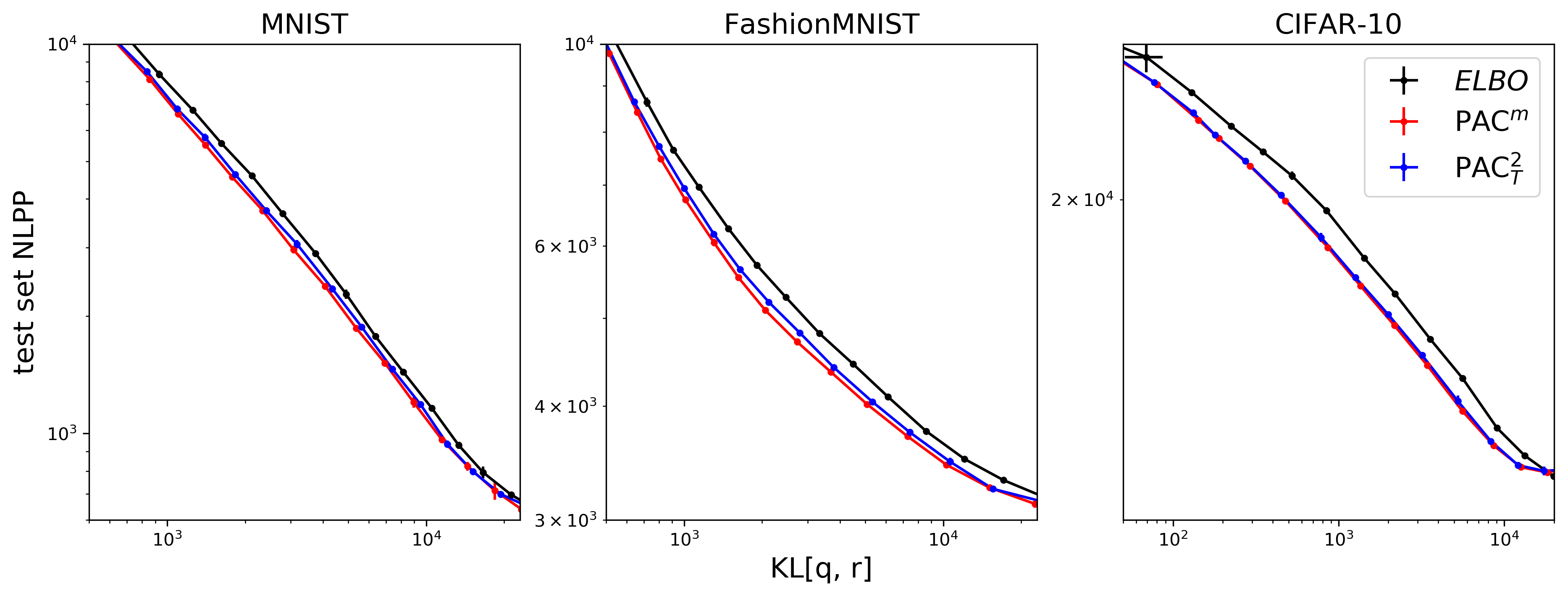}
    \caption{Predictive negative log-likelihood as a function of the $\KL$ divergence between the learned posterior and the prior (a measurement of the information content contained in the posterior distribution).  The Pareto frontier for each model is shown as the solid line, as measured by the indicated points.  Lower and to the left is ``better.'' While all models have comparable performance, we find that models which optimize \pac-Bayesian bounds on the predictive likelihood do a better job of distilling information from the dataset, and therefore require fewer bits to produce equivalent accuracies.  \pacm\ performs best.
    \label{fig:global_classification}}
\end{figure}

We experimented with classification using a Bayesian Neural Network on several popular benchmarking datasets.  Experimental details can be seen in \autoref{sec:extraclass}.  Similar to \cite{alemi2018fixing}, we evaluate our models as a function of the constant $\beta$ by producing the relationship between predictive negative log-likelihood (distortion) and $\KL$ divergence in the model (rate), a measure of compression of the model.  For global experiments, we show the results in Figure~\ref{fig:global_classification}.  At the end of the day, all models achieve a comparable accuracy (within the experimental uncertainty).  However, we find that \pacm and \pactwot\ do so at lower rate than \elbo, and also have the dominant Pareto-frontier in the information, indicating that it may be doing a better job of distilling useful information from the data.

\subsection{Bayesian Neural Network - Stochastic Activations}
\label{sec:activationexperiment}
We also consider classification using a different, and non-traditional, type of Bayesian Neural Network wherein we treat the activations of an intermediate layer in the model as the random variables.  This corresponds to the following graphical model.
\begin{align}
&\text{for } i = 1 \ldots n: \nonumber\\
&\quad  Z_i \sim r(Z) \\
&\quad  y_i \sim p(Y_i | Z_i)
\end{align}

In this formulation neither evidence $x_i$ nor deep neural network are directly present in the assumed generative process. Rather, these ideas appear only in the construction of the surrogate posterior, i.e., $Z_i \sim q(Z | x_i, \theta).$ For example, one might assume $q(Z_i|x_i,\theta) \equiv \operatorname{\sf Normal}(\mu_x, \sigma_x)$ where $\mu_x,\sigma_x$ are computed from two outputs of a DNN evaluated on $x_i$ and using parameters $\theta$ (both of which are regarded as being non-stochastic).  This type of setup is familiarized by Variational Autoencoders \cite{kingma2013auto}, and in deep variational information bottleneck \cite{alemi2016deep} models which use this graphical model to optimize for the log-evidence (or a bound on mutual informations) to set up either an unsupervised generative model (VAE) or a supervised predictive model (VIB).  For these experiments, we follow this previous work and use a deep neural network as an ``encoder'' which predicts the parameters of the posterior distribution, and a ``decoder'' which uses the latent variable to define $p(Y_i|Z_i)$.  As in \cite{kingma2013auto}, we use the \emph{reparameterization trick} to differentiate through the posterior sampling, which facilitates the optimization of the encoder parameters.

\begin{figure}[htbp]
    \centering
    \includegraphics[width=\linewidth]{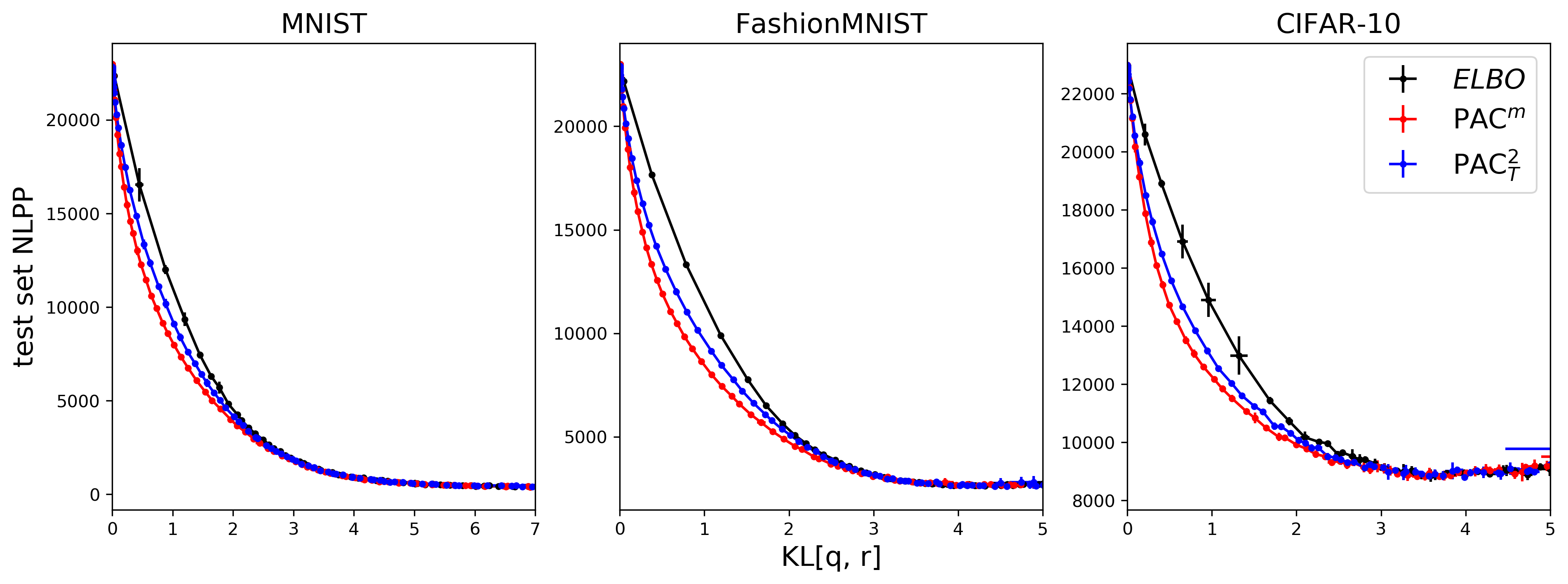}
    \caption{Similar to Figure~\ref{fig:global_classification}, but where the posterior is defined over activations of an intermediate layer of the network, rather than all of the weights.  Similar to before, lower and to the left is ``better.''  We find that in this context, \pacm\ clearly has the dominant pareto frontier.}
    \label{fig:local_classification}
\end{figure}

\begin{figure}[htbp]
    \centering
    \includegraphics[width=\hsize]{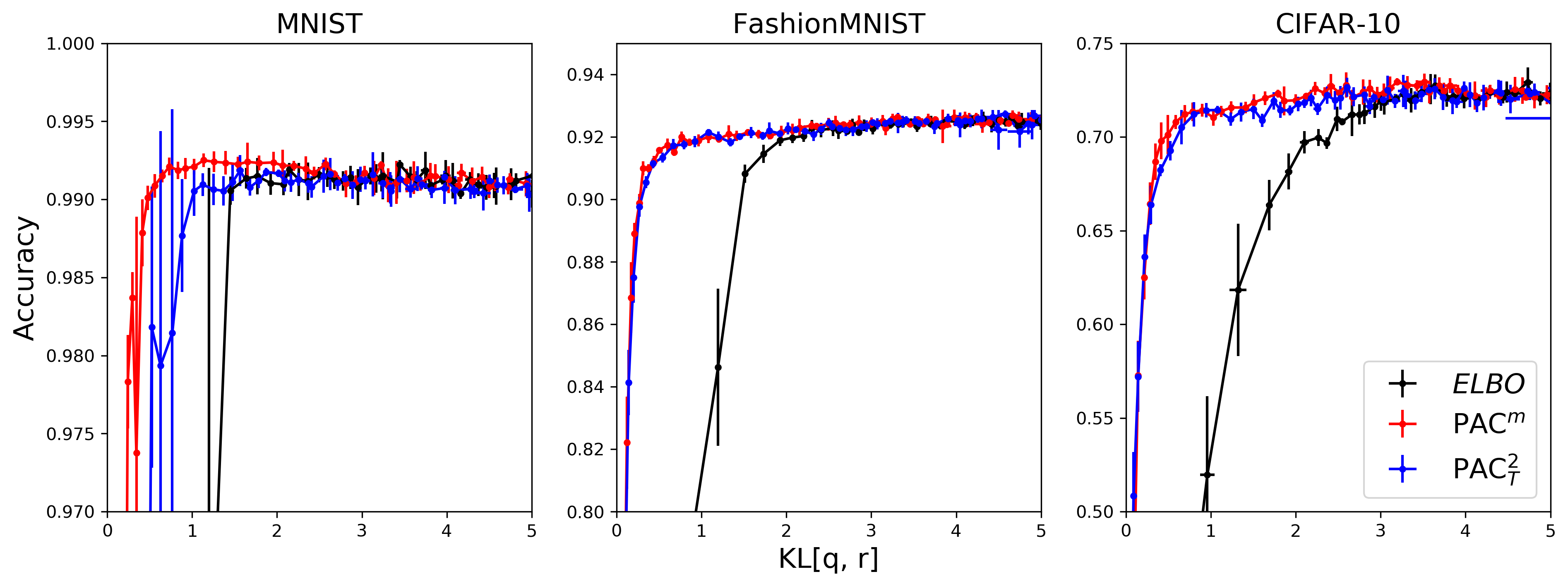}
    \caption{Classification accuracy as a function of the KL Divergence between the posterior and the prior.  \pacm\ and \pactwot\ consistently offer higher accuracy as a function of $\KL$ (i.e. for more \emph{compressed} posteriors), with \pacm\ appearing to do slightly better on MNIST.  Sharp decreases in classification accuracy, along with corresponding large uncertainty in final accuracy correspond to a sudden collapse of the posterior which occurs for sufficiently large $\beta$.}
    \label{fig:local_classification_acc}
\end{figure}

Experimental details are presented in \autoref{sec:extraclass}.  To evaluate performance, we show the Negative log-posterior-predictive probability as a function of the KL divergence between the posterior and the prior.  The results are shown in \autoref{fig:local_classification}.  Notably, we find that \pacm\ has a Pareto frontier which advances noticeably beyond the other alternatives.  This means that the model needs to learn less information in the posterior in order to make reasonable predictions on the data.  We further show the classification accuracy as a function of $\KL$ in Figure~\ref{fig:local_classification_acc}.  We find that both \pacm\ and \pactwot\ appear to require significantly less information in the latent representation in order to make useful predictions.  \pacm\ still appears to have the dominant Pareto frontier in this space, though it is often ambiguous that it performs ``better'' than \pactwot\ in this space.  However, it still appears that both outperform \elbo\ which appears to undergo posterior collapse at relative high rates, as indicated by the sudden sharp decrease in accuracy.

\section{Experimental Details}
\label{sec:experimentaldetails}

\subsection{Example Code}
\label{sec:examplecode}
Here we provide example code for computing each loss.  In all cases, we assume that one has a posterior, prior, and likelihood, where the posterior and prior are over the weights of the model, and the likelihood is a function which takes in weights and inputs and returns a probability distribution over outputs.  All of the following use tensorflow and tensorflow probability \cite{abadi2016tensorflow, dillon2017tensorflow}.  Additional arguments are $xy$ and $y$; the inputs to the model and outputs from the model, $m$; the number of samples to draw from the posterior, $\beta$; the weight to place on the $\KL$ penalty, and $n$; the number of examples in the dataset.

\begin{figure}[htbp]

\begin{lstlisting}[language=python]
def elbo(prior, likelihood, posterior, x, y, m, beta, n):
  w = posterior.sample(m)
  ll = likelihood(x, w).log_prob(y)
  kl = tf.reduce_mean(
      posterior.log_prob(w) - prior.log_prob(w),
      axis=0)
  nll = -tf.reduce_mean(ll, axis=(0, 1))
  return nll + kl / (beta * n)
\end{lstlisting}

\caption{\href{http://tensorflow.org/probability/}{TF Probability} \citep{dillon2017tensorflow} implementation of \elbo\ loss for a unimodal global latent variable models (e.g., BNN).}
\label{fig:elbo-code}
\end{figure}

\begin{figure}[htbp]
\begin{lstlisting}[language=python]
def pacm(prior, likelihood, posterior, x, y, m, beta, n):
  w = posterior.sample(m)
  ll = likelihood(x, w).log_prob(y)
  kl = tf.reduce_mean(
      posterior.log_prob(w) - prior.log_prob(w),
      axis=0)
  nlpp = -tf.reduce_mean(
      tfp.math.reduce_logmeanexp(ll, axis=0),
      axis=0)
  return nlpp + kl / (beta * n)
\end{lstlisting}
\caption{\href{http://tensorflow.org/probability/}{TF Probability} \citep{dillon2017tensorflow} implementation of \pacm\ loss for a unimodal global latent variable models (e.g., BNN).  Note that this is identical to \elbo, with the exception of the use of the negative log-posterior-predictive rather than the negative log-likelihood.}
\label{fig:pacm-code}
\end{figure}

Note that \pacm\ and \elbo\ are almost identical.  The only difference between the two is that \pacm\ uses a log-mean-exp over the sample dimensions to get the negative log-posterior-predictive probability rather than the expected negative log-likelihood.  Note also that this is not the case with \pactwot, which relies on the additional computation of a complicated variance term.  This term has memory and compute cost which scales in the number of samples, though this will likely be sub-dominant to the memory cost of the forward pass in the network itself.  It also relies on several tricks to encourage stability, and for the likelihood to be bounded in order for it to not converge to $-\infty$.

\begin{figure}[htbp]
\begin{lstlisting}[language=python]
def pac2t(prior, likelihood, posterior, x, y, m, beta, n, smoothing_constant=0.1):
  w = posterior.sample(m)
  ll = likelihood(x, w).log_prob(y)
  nll = -tf.reduce_mean(ll, axis=(0, 1))
  kl = tf.reduce_mean(posterior.log_prob(w) - prior.log_prob(w), axis=0)
  # We now compute the Masegosa "variance."
  lmx = tf.stop_gradient(
      tf.reduce_max(ll, axis=0, keepdims=True) + smoothing_constant)
  ll_max_centered = ll - lmx
  al = tfp.math.reduce_logmeanexp(ll_max_centered, axis=0)
  h = 2. * tf.stop_gradient(al / (1 - tf.math.exp(al))**2 +     
      1. / (tf.math.exp(al) * (1 - tf.math.exp(al))))
  var1 = h * tf.math.exp(2 * ll_mac_centered)
  var2 = tf.math.reduce_mean(
      h * tf.math.exp(
          ll_max_centered[tf.newaxis] + 
          ll_max_centered[:, tf.newaxis]),
      axis=0)
  variance = tf.math.reduce_mean(var1 - var2, axis=(0, 1))
  return nll - variance + kl / (beta * n)
\end{lstlisting}
\caption{\href{http://tensorflow.org/probability/}{TF Probability} \citep{dillon2017tensorflow} implementation of \pactwot\ loss for a unimodal global latent variable models (e.g., BNN).}
\label{fig:pac2t-code}
\end{figure}

\subsection{Toy Model}
\label{sec:toydetails}

The toy problem in \cref{fig:toy} was as described in \cref{sec:toy}. The true data distribution 
came from a 30-70 mixture of two \normal\ distributions, with a variance
of 1 and means at -2 and 2.  The model was a standard \normal\
with fixed unit variance, the only learned parameter being the mean.
Five datapoints were drawn, as indicated by the hash marks near the axis in the figures.  The inferential risks were determined analytically, as the solution takes on the closed form~\citep{murphy2007conjugate}.

For the \pac-predictive risk, the posterior was found numerically with
an iterative procedure.
The sought after parameter distribution was represented by 
the values the density took on a grid with 500 points from -30 to 30.
If we take $m\to \infty$ in $\pacpredrisk_{n,m}$ in \cref{eqn:pacpredrisk}, we have:
\begin{equation}
    \pacpredrisk_{n,\infty}[q; r, \beta] = -\frac 1 n \sum_i^n \log \left( \int d\theta \, q(\theta) p(x_i|\theta) \right) + \frac 1 {\beta n } \kl{q(\Theta)}{r(\Theta)} 
\end{equation}
Trying to minimize this functional with respect to $q(\Theta)$ using calculus of variations (along with the constraint that $q(\Theta)$ integrates to 1) suggests an iterative procedure to find the optimal parameter distribution:
\begin{align}
    q^{n+1}(\Theta) &\propto r(\Theta) \exp \left(\beta \sum_i \frac{p(x_i|\Theta)}{p^{(n)}(x_i)} \right) \label{eqn:iter1}\\
    p^{(n+1)}(x_i) &= \alpha p^{(n)}(x_i) + (1-\alpha) \int d\theta\, q^{(n+1)}(\theta) p(x_i|\theta). \label{eqn:iter2}
\end{align}
We iterated these equations numerically, representing the parameter
distribution as the values it took on a grid of 500 points between -30, and 30.  \cref{eqn:iter1} sets the new estimate for the parameter
distribution in terms of the current estimate for the data point
marginal likelihoods.  Notice the proportionality here, as we then
numerically normalized the density after setting it to the right 
hand side of \cref{eqn:iter1}.  Then in \cref{eqn:iter2} we update
our estimates of the data point marginal likelihoods, which act as 
sort of weights for the generalized Boltzmann distribution
that is our parameter distribution. For the figure in the paper
the mixing fraction $\alpha$ was set to $0.9$.

The empirical predictive risk was minimized numerically.
For the empirical predictive risk, an explicit mixture was fit, 
in this instance a 300 component \normal\ distribution, all with
fixed unit variance.  This is akin to searching for a 300 component
atomic posterior distribution ($q(\Theta) = \sum_i \lambda_i \delta(\Theta- \theta_i))$.  This was minimized with {\tt adagrad}
trained until it reached a fixed point to within a tolerance of $10^{-5}$.  Repeated runs all gave the same result.  Though this
was assuming the parameter distribution was itself atomic,
experiments with a setup as was done for the \pac-predictive risk
verified that the parameter distribution quickly does collect
to an delta-comb.

\subsection{Sinusoid}
\label{sec:sinusoiddetails}
As mentioned in \cref{sec:sinusoid}, for our first experiment we tried to predict data drawn from the following sinusoid model:
\begin{align}
&\text{for } i = 1\ldots n: \nonumber\\
&\quad \mu_{x_i} = 7\sin{\left(\frac{3x_i}{4}\right)}+\frac{x_i}{2} \\
&\quad Y_i \sim \operatorname{\sf Normal}(\mu_{x_i}, 10).
\end{align}
We generate data for $10^4$ evenly spaced values of $x\in \left[-10.5, 10.5\right]$.

For our neural network, we use a two layer \mlp\ with 20 hidden units, and a hyperbolic tangent activation function.  We use a \normal\ distribution for the posterior, with both mean and variance as trainable variables.  The initial values of the means were set to 0, and the initial variances were set to 1.  The variances were constrained to be positive using the $\exp$ bijector available in tensorflow probability.  We similarly use $\operatorname{\sf Normal}(0,1)$ for the prior over each weight and bias.  For the likelihood, we use the MLP to predict the mean of a normal distribution, whose variance is fixed to 1.  

We train all models using Adam \citep{kingma2014adam} with a learning rate of 0.01 and no learning rate decay.  We use full batch training, for $10^5$ steps.  For \cref{fig:toy_experiments}, we used $m=100$ samples from the posterior during training, and $\beta=1$ for all models.  For all models, we evaluate performance using the log-posterior predictive, constructed using $10^3$ samples from the posterior.  

\subsection{Mixture Experiments}
\label{sec:mixturedetails}

For our second experiment, we use data generated from a two component mixture distribution:
\begin{align}
&\text{for } i = 1 \ldots n: \nonumber\\
&\quad \mu_{x_i} = 7\sin{\left(\frac{3x_i}{4}\right)}+\frac{x_i}{2} \\
&\quad Z_i \sim \operatorname{\sf Rademacher} \\
&\quad Y_i \sim \operatorname{\sf Normal}(Z_i\mu_{x_i}, 1)
\end{align}

The model setup was largely similar to the Sinusoid experiment described in \cref{sec:sinusoiddetails}, but with one major difference:  For these experiments we added an additional hidden layer to the networks to aid in expressiveness.  We also used Exponential Linear Unit ({\sf ELU}) activations instead of {\sf tanh} to facilitate gradient propagation more easily.  For these experiments we used both a unimodal posterior, as well as a mixture posterior, but the underlying distribution was implemented similarly to the sinusoid (i.e. a $\operatorname{\sf Normal}$ distribution with learnable mean and variance).  When considering a multimodal posterior, we fixed the component probabilities to 0.5 and used stratified sampling to integrate over the discrete categorical random variable.  For unimodal likelihoods, we used a normal distribution whose mean was predicted by the model, and which had a fixed variance of 1.  When considering a mixture likelihood, we compared situations with both 1 and 2 components in the posterior.  The MLP was set up to predict the means of a two component mixture of Gaussian distributions, whose component probabilities were fixed to 0.5, and whose component variances were fixed to 1.  

All models were again trained using Adam with an initial learning rate of 0.01, but this time we added a small amount of learning rate decay with a decay rate of 0.5 and a decay timescale of $10^5$ steps.  Because the model was only trained for $10^5$ steps, the learning rate only undergoes one half-life.  We did not study if holding the learning rate fixed changed the results at all, though it is doubtful that it did.  We employed full batch training, and used $m=100$ samples from the posterior.  To evaluate the models qualitatively, as in \cref{fig:masegosa_mixture_1comp}, we used $10^5$ samples from the posterior to construct the predictive distribution.  For each sample, we computed a forward pass for $10^3$ evenly spaced values of $x$ and drew a single sample from the resulting likelihood.  This gave us $10^{5}$ samples from the predictive distribution for each $x$.  We then computed the 1-d histogram of the predictive distribution at each $x$, which we used to display the predictive models as in 
\cref{fig:toy_experiments}.  To quantitatively evaluate models, we used $10^4$ samples from the posterior to construct the predictive distribution, and then computed the $\KL$ divergence from the known generative distribution for each of the ($x$, $y$) pairs in an independently generated test set.

\subsection{Image Experiments}
\label{sec:extraclass}

\subsubsection{Structured Prediction}
For the structured prediction experiments, we attempt to solve the following problem:  Given the top half of an image, predict the bottom half of the image. We follow the setup in \citet{masegosa2019learning}, for this experiment which we now describe.  We use the experimental TFP Neural Networking toolbox (\verb|tfp.experimental.nn|, \citep{dillon2017tensorflow}) and TFP Joint Distributions \citep{piponi2020joint} to compactly specify BNNs.  For the posterior and prior, we used Normal distributions.  The posterior had learnable variables to represent the mean and variance, while the mean and variance in the prior are assumed to be constant.  The model predicts the means of independent Normal distributions with fixed variance of $1/255$ which we use as our likelihood for each pixel.  We assume that all pixels are independent and therefore ignore covariance between pixels in the output distribution.  Note that this assumption is (purposefully) incorrect; images generally have strong correlations between adjacent spatial pixels and between the colors within a single pixel.  For the network architecture, we used a 3-layer MLP with 50 hidden units, and {\sf ELU} activations.  We therefore fed our input images to the model as flattened vectors.  For CIFAR-10, we converted the image to grayscale to reduce the number of pixels and simplify the model.

All models were trained using Adam with an initial learning rate of 0.001, decayed by 0.5 every $10^5$ steps.  We train models for 500 epochs.  We used a batch size of 128 during training.  We tested performance on the heldout evaluation set as a function of $m$, ranging from $m=1$ to $m=32$, where each $m$ corresponds to the number of samples used during training.  Reconstruction performance was quantified using the log posterior-predictive ($\lpp$), which we measure using 100 samples from the posterior. We train 5 different models independently, with different random initializations and different shufflings of the training set in order to obtain the uncertainties in the final performance of the model. 

\subsubsection{Classification - Stochastic Weights}

For our likelihood, we use a categorical distribution with 10 possible outcomes (all image datasets we consider have 10 output classes).  Similar to previous experiments, we used Normal distributions for both the Posterior and the Prior, where the posterior uses variables to represent the location and scale of the normal distribution, and where the prior uses fixed values, both of which were initialized or fixed to 0 for the location and 1 for the scale respectively.  We used the same architecture as the structured prediction experiments:  A 3-layer MLP with 50 hidden units and {\sf ELU} activations.  All input images were normalized to the [-1, 1] interval before being passed to the first layer of the network.

Similar to the structured prediction experiments, we trained with a batch size of 128.  We optimized our model for 100 epochs using Adam, with an initial learning rate of $10^{-4}$, which we decayed by a factor of 0.5 every $10^5$ steps.  We study performance as a function of $\beta$, and fix the number of samples used during training to $m=4$.  We consider 33 values of $c$ spaced logarithmically and ranging from $\left[10^{-3}-10^{3}\right]$.  We evaluate performance by computing both the log-posterior-predictive $\lpp$, on the heldout evaluation set.  We use 100 samples to construct the posterior predictive distribution.

\subsubsection{Classification - Stochastic Activations}

For this model, we use the latent embedding $z_i$ to codify each example in a 16 dimensional latent space.  To increase the capacity of the model and since the memory cost is much lower, we consider a convolutional neural network for the encoder.  This layer uses the following architecture\footnote{This architecture follows the encoder from the VAE example at
\url{https://www.tensorflow.org/probability/examples/Probabilistic_Layers_VAE}
}: 
4 convolutional layers, followed by 2 dense layers.  For each layer, we use {\sf LeakyReLU} activations.  Alternating convolutional layers use a stride of 2.  The last dense layer predicts the parameters of the posterior.  All images were normalized to the range [-1, 1] prior to being passed to the network.

For the posterior, we used a Multivariate Normal Distribution. The encoder predicts the location and the cholesky decomposition of the precision matrix.  For the prior, we used a Multivariate Normal distribution with mean 0, and an identity covariance.  For the likelihood, we used a categorical distribution.  During training, we used a batch size of 128.  We optimized our model for 100 epochs, using Adam with an initial learning rate of $10^{-4}$, which was decayed by a factor of 0.5 every $10^5$ training steps.  Similar to our stochastic weights experiments, we evaluated performance as a function of $\beta$ using $100$ log-spaced bins between $10^{-3}$ and 10.  We evaluate performance by computing both the $\lpp$ and the Accuracy on the evaluation set.  For both, we used 100 samples from the posterior to construct the predictive distribution.

\end{document}